\documentclass{article}

\usepackage{graphics} 
\usepackage{epsfig} 
\usepackage{times} 
\usepackage{amsmath} 
\usepackage{amssymb}  
\usepackage{threeparttable}
\usepackage{multirow}
\usepackage{algorithm}
\usepackage{algpseudocode}
\usepackage{booktabs}
\usepackage{pifont}
\usepackage{xspace}
\usepackage{adjustbox}
\usepackage{caption}
\usepackage{enumitem}
\usepackage[table]{xcolor}

\newcommand{\cmark}{\ding{51}}
\newcommand{\xmark}{\ding{55}}
\newcommand{\ourmethod}{IntentNav\xspace}

\usepackage[final]{corl_2026} 

\makeatletter
\renewcommand{\section}{%
  \@startsection{section}{1}{\z@}%
                {-1.3ex \@plus -0.5ex \@minus -0.2ex}%
                { 0.8ex \@plus  0.3ex \@minus  0.2ex}%
                {\large\bf\raggedright}%
}
\makeatother

\title{IntentNav: Learning Spatial-Visual Object Navigation from Human Demonstrations}

\author{
\textbf{Yuxin Cai}$^{1,2,4,*}$,
\textbf{Zongtai Li}$^{2,*}$,\allowbreak
\textbf{Maonan Wang}$^{3}$,\allowbreak
\textbf{Muyi Bao}$^{2}$,\allowbreak
\textbf{Haokun Zhu}$^{2}$,\allowbreak
\textbf{Ruofei Bai}$^{1,4}$\\
\textbf{Ding Zhao}$^{2}$,
\textbf{Zirui Li}$^{1}$,
\textbf{Wenshan Wang}$^{2}$,
\textbf{Wei-Yun Yau}$^{4}$,
\textbf{Ji Zhang}$^{2}$,
\textbf{Chen Lv}$^{1,\dagger}$\\
$^{1}$ Nanyang Technological University \quad
$^{3}$ The Chinese University of Hong Kong\\
$^{2}$ Carnegie Mellon University \quad
$^{4}$ A*STAR Institute for Infocomm Research (I2R)\\
$^{*}$ Equal contribution.
$^{\dagger}$ Corresponding author.
}

\begin{document}
\maketitle

\vspace{-2.0em}
\begin{figure}[h]
    \centering
    \includegraphics[width=0.95\linewidth]{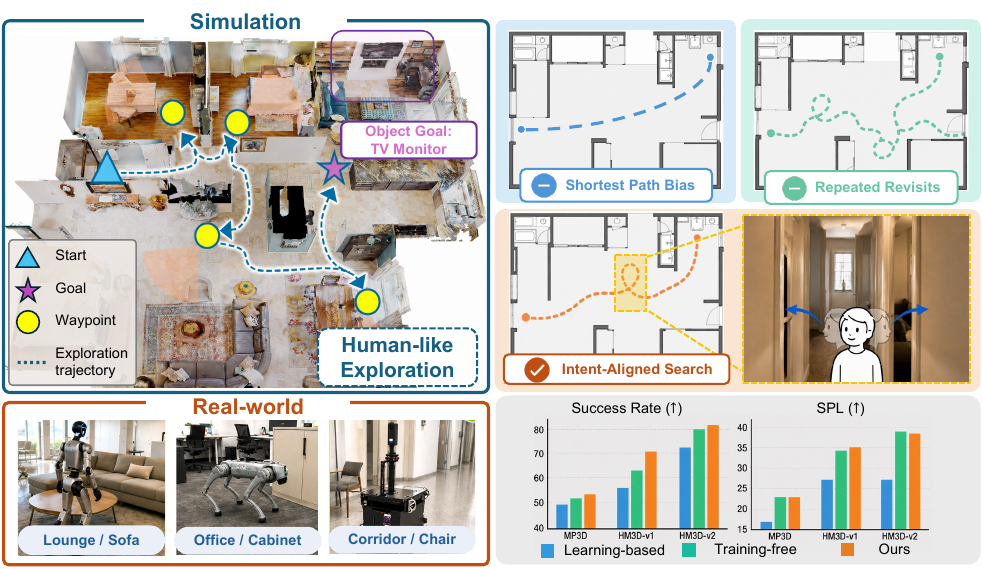}
    \caption{
    \textbf{IntentNav learns spatial-visual ObjectNav from human demonstrations.}
    It grounds frontier and target decisions in a unified BEV space, yielding directed search behavior, strong benchmark performance, and transfer across wheeled, quadruped, and humanoid robots.
    }
    \vspace{-1.0em}
    \label{fig:teaser}
\end{figure}

\begin{abstract}
Object navigation requires a robot to search for an unobserved target in an unknown environment by deciding where to explore next under partial observability.
Effective search resembles human-like exploration: selectively probing visually promising frontiers while relying on spatial memory to avoid redundant revisits.
We propose \textbf{\ourmethod}, a spatial-visual imitation framework that learns human-like ObjectNav policies from human demonstrations.
To infer high-level search intent from low-level human actions, we introduce Frontier-based Human-Intent Labeling, which looks ahead in human demonstrations and labels the frontier that best explains the demonstrator's future search direction.
We construct a spatial-visual candidate space, where BEV memory tracks explored regions, unexplored frontiers, and trajectory history, while egocentric visual memory provides semantic cues for each candidate.
A VLM policy is trained to select among these grounded candidates, using Intent-Aligned Objective to encourage consistent and human-like exploration.
\ourmethod achieves state-of-the-art performance among learning-based methods on the MP3D, HM3D-v1, and HM3D-v2 ObjectNav benchmarks, attaining the best success rate among all compared methods.
The proposed candidate-level navigation interface transfers zero-shot to wheeled, quadruped, and humanoid robots without further VLM fine-tuning.
\href{https://wmn7.github.io/IntentNav/}{Project page}.
\end{abstract}


\section{Introduction}
\vspace{-0.3em}
\label{sec:intro}
Object Navigation (ObjectNav) requires an embodied agent to search for and navigate to an instance of the target object category in an unseen environment under partial observability~\citep{batra2020objectnav, anderson2018evaluation, sun2024survey}.
The task has become increasingly important, as locating everyday objects is essential for robots operating in homes and other human-centered spaces.
The central challenge is to perform long-horizon, target-conditioned search before the target is observed, inferring promising search directions from partial observations and continually adapting as new evidence appears.

Humans are naturally adept at this form of search.
Imagine entering an unfamiliar apartment and being asked to find a mug.
Rather than exhaustively traversing every reachable area, people use spatial and visual cues to guide the search: moving toward regions resembling kitchens or dining areas, pausing at informative viewpoints to scan the surroundings, and briefly probing uncertain rooms or hallways before deciding whether to continue.
Their strong spatial awareness enables them to keep track of where they have been, which regions remain unexplored, and how the environment is laid out.
Semantic common sense further helps them infer where the target object is likely to appear, allowing promising areas to be inspected carefully while unlikely ones are passed over quickly.

Recent advances in vision-language models (VLMs) have brought strong semantic understanding and commonsense reasoning capabilities to ObjectNav.
Recent works~\citep{zhou2023escexplorationsoftcommonsense, yokoyama2024vlfm, kuang2024openfmnav, yin2024sg, cao2025cognavcognitiveprocessmodeling, zhu2025strive} leverage VLMs to decide where to explore by reasoning about target relevance from visual and spatial context.
However, VLM reasoning alone does not guarantee stable long-horizon, target-conditioned search behavior.
VLMs still have limited spatiotemporal understanding~\citep{zhang2024dovision, chen2024spatialvlm}, and existing VLM-based policies often rely on first-person RGB observations~\citep{tsai2023multimodallargelanguagemodel, li2025compassnav, wang2026hydra, zhang2024navidvideobasedvlmplans, zhang2024uni, xue2025omninav} or text-based map summaries~\citep{zhu2025strive, zhu2026sysnav, wu2024voronav, yin2024sg, cao2025cognavcognitiveprocessmodeling}.
Such representations abstract away geometric details and fine-grained visual context, making it difficult to maintain a structured spatial-visual memory over long horizons.
For methods that predict low-level actions~\citep{majumdar2022zson, zhang2024navidvideobasedvlmplans, zhang2024uni, wang2026hydra}, search decisions can further become overly short-horizon and reactive.
As a result, these policies may produce locally plausible decisions at individual steps, yet fail to maintain a coherent strategy over time, leading to repetitive behaviors such as turning in place, backtracking, or repeatedly inspecting the same region.

We introduce \textbf{\ourmethod}, a BEV-grounded spatial-visual imitation framework that trains a VLM policy to learn candidate-level ObjectNav decisions from human demonstrations.
The central idea is to infer the underlying high-level search intent from low-level human actions and express it as a candidate-level waypoint decision in a unified BEV space.
We address this with Frontier-based Human-Intent Labeling, which looks ahead into the demonstration and uses frontier evolution to identify the current frontier that best explains the demonstrator's future search direction.
The shared BEV space provides a persistent spatial frame for scene layout, explored history, frontier regions, and provisional target detections, providing the structured spatial context needed for long-horizon search.
Rather than predicting low-level actions, the policy selects among reachable waypoint candidates defined in this shared space.
Each candidate is grounded by its first-observed egocentric visual context and agent-centric relative geometry, enabling joint reasoning over semantic evidence, spatial relations, and search history.
We train this candidate-level policy with Intent-Aligned Objective, encouraging directionally consistent exploration and decisive target commitment.

To summarize our contributions:
(1) We introduce Frontier-based Human-Intent Labeling, which derives candidate-level proxy supervision from low-level human trajectories, enabling learning of human search intent.
(2) We propose a BEV-grounded spatial-visual candidate policy that unifies exploration and target commitment within a shared decision space.
(3) \ourmethod IntentNav achieves state of-the-art performance on MP3D, HM3D-v1, and HM3D-v2 Object Navigation benchmarks, and transfers zero-shot across wheeled, quadruped, and humanoid robots.

\section{Related Work}
\vspace{-0.3em}
\label{sec:related}
Recent ObjectNav methods increasingly leverage pretrained multimodal models to guide long-horizon target search. These methods can be broadly grouped by their decision interface: map-mediated reasoning or egocentric policy learning.

\paragraph{Map-mediated reasoning.}

One line of VLM-based ObjectNav methods organizes accumulated observations into explicit map representations and performs navigation reasoning over the resulting map structures.
Some approaches maintain dense spatial memories, such as BEV semantic maps~\citep{yu2023l3mvn, zhang2025apexnav, cao2025cognavcognitiveprocessmodeling} or language-aligned value maps~\citep{yokoyama2024vlfm, zhong2025topvnavunlockingtopviewspatial, kuang2024openfmnav}, where VLMs mainly provide semantic priors for scoring directions, frontiers, or candidate goals.
Other works further abstract environments into structured topological representations, including Voronoi graphs~\citep{wu2024voronav} and scene graphs~\citep{yin2024sg, zhu2025strive, zhu2026sysnav}.
Since map representations are not native inputs to pretrained VLMs, these methods typically summarize map-level spatial context into textual descriptions for VLM reasoning, often losing spatial structure and fine-grained visual cues.
In contrast, \ourmethod learns candidate-level search decisions in a global BEV decision space that preserves candidate-specific visual evidence and geometry.

\paragraph{Egocentric policy learning.}

Another line of work fine-tunes VLMs on navigation data, typically predicting low-level actions from first-person images, videos, or observation histories~\citep{tsai2023multimodallargelanguagemodel, li2025compassnav, wang2026hydra, zhang2024navidvideobasedvlmplans, zhang2024uni, xue2025omninav}.
These methods preserve rich visual observations and enable end-to-end language-conditioned navigation policies.
However, first-person RGB inputs lack explicit spatial structure, and low-level action prediction emphasizes short-horizon control, often leading to locally repetitive behaviors, such as turning in place or repeatedly backtracking through identical regions.
Moreover, these methods are commonly trained on oracle shortest-path trajectories~\citep{wang2026hydra, zhang2024uni, li2025compassnav}, sometimes augmented with exploration data~\citep{xue2025omninav}.
This supervision primarily optimizes goal reaching under full observability, rather than modeling high-level human search intent under partial observability.
\ourmethod instead learns waypoint-level decisions from human demonstrations in a spatially grounded BEV decision space, directly modeling human-like, target-conditioned search behavior under partial observability and supporting more effective long-horizon exploration.

\section{Method}
\label{sec:method}

\subsection{Candidate-Level ObjectNav Formulation}
\label{sec:method-formulation}

\begin{figure}[t]
    \centering
    \includegraphics[width=1.0\linewidth]{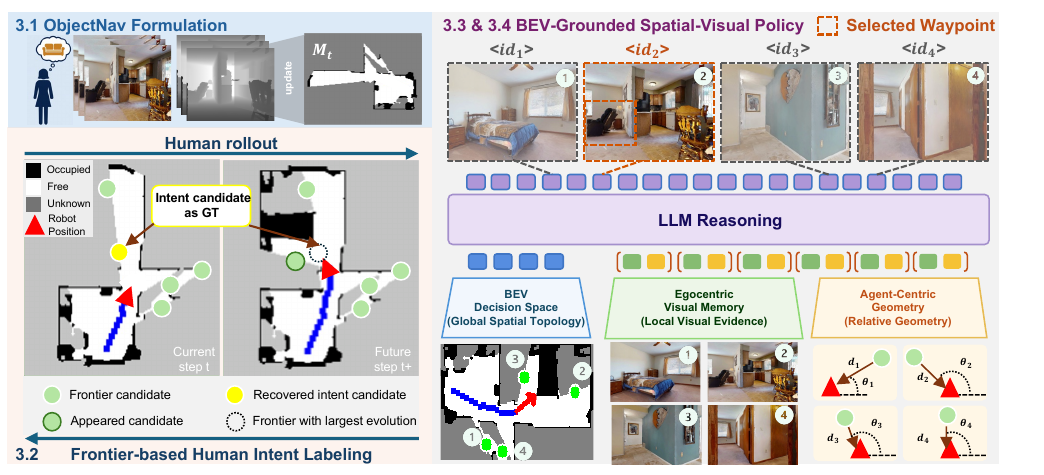}
    \caption{
    \textbf{\ourmethod overview.}
    \textbf{Left:} ObjectNav is formulated as BEV candidate-level waypoint selection, with Frontier-based Human-Intent Labeling recovering supervision from low-level human rollouts.
    \textbf{Right:} The VLM policy combines the BEV decision space, candidate-aligned visual memories, and agent-centric geometry, then predicts a reserved candidate-ID token as the next waypoint.
    }
    \label{fig:architecture}
\end{figure}

As illustrated in Fig.~\ref{fig:architecture}, at each decision timestep $t$, the agent receives RGB-D observations and updates a persistent BEV map $M_t$ that records explored free space, obstacles, trajectory history, and unknown regions.
From $M_t$, we construct a unified BEV candidate set $\mathcal{C}_t=\{x_i\}_{i=1}^{n_t}$ containing frontier candidates and target candidates.
Frontier candidates lie on navigable boundaries between explored and unknown regions, while target candidates correspond to verified target hypotheses.
Each candidate $x_i\in\mathbb{R}^2$ denotes a location in the BEV space and is paired with a candidate-specific egocentric RGB image $I_i^{\mathrm{ego}}$ and an agent-centric relative-geometry feature $\mathbf{f}_i$.

Given the target category $c$, the VLM predicts a candidate index over the current candidate set by:
\begin{equation}
\hat{k}
=
\arg\max_{k\in[n_t]}
p_\theta\!\left(
k \mid M_t,\{x_i,I_i^{\mathrm{ego}},\mathbf{f}_i\}_{i=1}^{n_t},c
\right),
\qquad
\hat{x}_t=x_{\hat{k}} .
\label{eq:policy}
\end{equation}
The selected waypoint $\hat{x}_t$ is handed to a platform-specific local planner and controller for execution, decoupling high-level decision making from embodiment-specific control and enabling transfer across different physical embodiments. Real-world deployment details are provided in Appendix~\ref{app:physical-deployment}.

\subsection{Frontier-based Human-Intent Labeling}
\label{sec:method-frontier-collapse}

Human demonstrations contain rich object-search intent but provide only low-level action sequences instead of waypoint-level decisions.
To bridge this gap, we replay Habitat-Web ObjectNav human demonstrations~\citep{ramrakhya2022habitat} on MP3D scenes~\citep{chang2017matterport3d} within the Habitat simulator~\citep{puig2023habitat} to construct candidate-level supervision.
If a target candidate is present in the current candidate set, we directly use its index as supervision.
Otherwise, we infer the human's exploration intent from future frontier evolution.

Let $\mathcal{F}_t = \{f_i^t\}$ denote the frontier set at step $t$, and define the point-to-set distance $d(x, \mathcal{F}) = \min_{f\in \mathcal{F}} \mathrm{geo}(x, f)$, where $\mathrm{geo}(\cdot,\cdot)$ denotes geodesic distance over navigable free space.
Starting from step $t$, we ignore short local adjustments by advancing to step $t_{\mathrm{adv}}$, after the demonstrator has executed at least six \texttt{move-forward} actions.
We then identify the earliest subsequent step $t^+$ where at least one current frontier no longer aligns with the future frontier set:
\begin{equation}
t^+ = \min \left\{ t' \ge t_{\text{adv}} \;\middle|\; \exists f \in \mathcal{F}_t, d(f, \mathcal{F}_{t'}) > \sigma_{\mathrm{evo}} \right\}, \quad \sigma_{\mathrm{evo}} \ \text{\small: frontier-evolution threshold}.
\end{equation}
The intent label $y_t$ is then assigned over the current candidate set $\mathcal{C}_t = \{x_i\}$:
\begin{equation}
y_t = \arg\max_{i} d(x_i, \mathcal{F}_{t^+}).
\label{eq:collapse}
\end{equation}
A large distance $d(x_i,\mathcal{F}_{t^+})$ indicates that candidate $x_i$ is most inconsistent with the future frontier set, suggesting that the demonstrator continued exploration through that region.
This converts low-level human trajectories into candidate-level proxy waypoint labels for training the VLM policy.

\subsection{BEV-Grounded Spatial-Visual Candidate Policy}
\label{sec:method-candidate-policy}

When searching for objects, humans combine spatial memory with local visual evidence.
We therefore represent each candidate with two complementary forms of grounding: an egocentric visual memory that captures what was observed around the candidate, and an agent-centric geometry feature that binds the visual evidence to the candidate's location in the BEV decision space.
Details of BEV construction, frontier extraction, and target verification are provided in Appendix~\ref{app:bev-map} and~\ref{app:target-verification}.

\paragraph{Candidate visual memory.}
To ground each BEV candidate $x_i$ with localized semantic evidence, it is paired with the egocentric RGB image $I_i^{\mathrm{ego}}$ captured when $x_i$ was first instantiated.
This memory augments the BEV waypoint with local visual context, such as room type, nearby objects, and scene layout, allowing the policy to assess the candidate's relevance to the target category.
Candidate IDs are defined within the current step candidate set rather than tracked persistently across timesteps, since frontier proposals can move, merge, or split as the map evolves.

\paragraph{Agent-centric geometry.}
Although the BEV map contains each candidate's spatial relation to the agent, this geometry may not be reliably extracted by the VLM from the map feature alone.
We therefore explicitly augment each candidate with an agent-centric relative-geometry feature.
For a candidate at map position $\mathbf{p}_i$, we compute this feature from the agent pose $(\mathbf{p}_a,\psi_a)$ via
\begin{equation}
\mathbf{f}_i =
\left[
\frac{\|\mathbf{p}_i-\mathbf{p}_a\|}{W},
\sin\theta_i^{\mathrm{rel}},
\cos\theta_i^{\mathrm{rel}}
\right],
\quad
\theta_i^{\mathrm{rel}}
=
\mathrm{wrap}\!\left(\mathrm{atan2}(-\Delta y,\Delta x)-\psi_a\right),
\label{eq:rel-feat}
\end{equation}
where $(\Delta x,\Delta y)=\mathbf{p}_i-\mathbf{p}_a$ and $W$ is the local BEV crop width.
This feature encodes the candidate's normalized distance and relative bearing from the agent.

\paragraph{Candidate-Aligned VLM Input.}
The VLM receives a shared BEV crop followed by serialized candidate blocks, each containing the candidate's visual memory and agent-centric geometry feature.
This organization preserves a one-to-one correspondence between each candidate, its local semantic evidence, and its explicit spatial grounding.
The BEV crop and candidate memory images are processed by separate vision encoders to handle the distinct visual statistics of top-down maps and egocentric views.
Prompt and model details are provided in Appendix~\ref{app:prompt} and ~\ref{app:architecture}.

\paragraph{Reserved Candidate-ID Selection.}
We reserve $32$ candidate-ID tokens in the model vocabulary, covering the maximum number of candidates used in our experiments.
In practice, each decision step contains approximately 18 candidates on average and at most 30 after filtering, so the reserved budget comfortably covers all active candidates (Appendix~\ref{app:training-corpus}).
The language model predicts a distribution over the reserved candidate-ID tokens.
This converts waypoint prediction into controlled candidate classification, avoiding free-form generation of indices.

\subsection{Intent-Aligned Objective}
\label{sec:method-intent-aligned}

Human search behavior often reflects directional intent rather than a commitment to one exact waypoint.
Nearby candidates along the demonstrated direction should be treated as plausible alternatives, but standard cross-entropy penalizes all incorrect candidates equally.

We therefore construct an angular soft target over $\mathcal{C}_t$.
Let $\theta_i$ and $\theta_{y_t}$ be the bearings from the agent to candidate $x_i$ and the labeled candidate $x_{y_t}$, respectively, and define $\Delta\theta_i=\mathrm{wrap}(\theta_i-\theta_{y_t})$.
The soft target is
\begin{equation}
q_i=
\frac{\exp\left(\kappa(\cos\Delta\theta_i-1)\right)}
{\sum_j\exp\left(\kappa(\cos\Delta\theta_j-1)\right)},
\qquad
\kappa=\sigma_{\mathrm{rad}}^{-2},
\label{eq:angular-target}
\end{equation}
with $\sigma_{\mathrm{rad}}=25^\circ\approx0.436\,\mathrm{rad}$.
This target assigns partial credit to candidates near the demonstrated direction while preserving strong penalties for large directional errors.
If the label is a frontier candidate, we combine hard candidate supervision with the angular soft target:
\begin{equation}
\mathcal{L}
=
(1-\lambda)\big[-\log p_\theta(y_t)\big]
+
\lambda\left[-\sum_{i\in[n_t]}q_i\log p_\theta(i)\right].
\label{eq:final-loss}
\end{equation}
For target candidate, we use hard cross-entropy only to encourage decisive commitment to the target.

\section{Experiments}
\label{sec:experiments}
\subsection{Experimental Setup}
\label{sec:exp-setup}

\paragraph{Training data and implementation.}
We replay Habitat-Web ObjectNav human demonstrations~\citep{ramrakhya2022habitat} on MP3D scenes~\citep{chang2017matterport3d} in Habitat~\citep{puig2023habitat}, and apply Frontier-based Human Intent Labeling to convert low-level action sequences into candidate-level waypoint labels.
The replayed corpus contains 28 scenes, 21 target categories, and 23,767 human demonstration episodes.
After filtering invalid replay states and episodes, the resulting dataset contains 2,363,108 candidate-level training samples.
\ourmethod is fine-tuned from InternVL3-2B~\citep{chen2025internvl3} using LoRA for 3 epochs on 4 A100 GPUs.
All ablations follow the same training protocol unless otherwise specified, with additional implementation details provided in Appendix~\ref{app:hyperparameters}.

\paragraph{Benchmarks and evaluation protocol.}
We evaluate \ourmethod on three Habitat ObjectNav validation benchmarks: MP3D~\citep{chang2017matterport3d, batra2020objectnav}, HM3D-v1~\citep{habitatchallenge2022}, and HM3D-v2~\citep{yadav2023habitat}.
Training demonstrations are replayed only on MP3D-train scenes, while MP3D-Val is used for in-domain evaluation.
Evaluation on HM3D measures zero-shot transfer to unseen scene domains.
We report Success Rate (SR), the proportion of episodes in which the agent reaches the target object within a success distance, and Success weighted by Path Length (SPL)~\citep{anderson2018evaluation}, which additionally accounts for path efficiency.

We further report four trajectory-level search metrics: Exploration Coverage Ratio (ECR) measures the fraction of reachable free space observed before termination; Local Scanning Ratio (LSR) measures in-place viewpoint scanning; Probing Motion Ratio (PMR) measures short out-and-back exploratory motions; and Redundant Revisit Ratio (RRR) measures revisits that do not expand explored area. Formal definitions and discussions are provided in Appendix~\ref{app:behavior-analysis}.

\subsection{Main Benchmark Results}
\label{sec:main-results}

\begin{table*}[t]
\centering
\small
\caption{
\textbf{Main ObjectNav benchmark results.}
Methods are grouped by training regime: training-free VLM systems, learning-based policies, generic-navigation models, and ours.
Best results are in \textbf{bold}, and second-best results are \underline{underlined}.
* indicates reproduced results.
}
\label{tab:train_results}
\begin{adjustbox}{center}
\begin{tabular}{lccccccc}
\toprule
Method & Learning
& \multicolumn{2}{c}{\textbf{MP3D}}
& \multicolumn{2}{c}{\textbf{HM3D-v1}}
& \multicolumn{2}{c}{\textbf{HM3D-v2}}
\\
\cmidrule(lr){3-4}
\cmidrule(lr){5-6}
\cmidrule(lr){7-8}
&
& SR (\%) $\uparrow$ & SPL (\%) $\uparrow$
& SR (\%) $\uparrow$ & SPL (\%) $\uparrow$
& SR (\%) $\uparrow$ & SPL (\%) $\uparrow$
\\
\midrule
VLFM~\cite{yokoyama2024vlfm}
& \xmark & 36.4 & 17.5 & 52.5 & 30.4 & 63.5 & 32.5 \\
SG-Nav-GPT~\cite{yin2024sg}
& \xmark & 40.2 & 16.0 & 54.0 & 24.9 & 49.6 & 25.5 \\
OpenFMNav~\cite{kuang2024openfmnav}
& \xmark & 37.2 & 15.7 & 54.9 & 24.4 & -    & -    \\
TriHelper~\cite{zhang2024trihelperzeroshotobjectnavigation}
& \xmark & -    & -    & 56.5 & 25.3 & -    & -    \\
BeliefMapNav~\cite{zhou2025beliefmapnav}
& \xmark & 37.3 & 17.6 & 61.4 & 30.4 & -    & -    \\
ApexNav~\cite{zhang2025apexnav}
& \xmark & 39.2 & 17.8 & 59.6 & 33.0 & 76.2 & 38.0 \\
WMNav~\cite{nie2025wmnavintegratingvisionlanguagemodels}
& \xmark & 45.4 & 17.2 & 58.1 & 31.2 & 72.2 & 33.3 \\
PIGEON~\cite{peng2025pigeon}
& \xmark & 41.6 & 14.4 & 57.9 & 32.3 & 79.2 & 36.8 \\
STRIVE~\cite{zhu2025strive}
& \xmark & \underline{52.3} & \underline{23.1} & 62.9 & \underline{34.2} & 79.6 & \textbf{38.7} \\
SysNav~\cite{zhu2026sysnav}
& \xmark & 50.7 & 18.1 & \underline{63.7} & 30.5 & \underline{80.8} & 37.2 \\

\midrule
DD-PPO~\cite{wijmans2019dd}
& \cmark & 21.6 & 8.7  & -    & -    & 27.9 & 14.2 \\
Habitat-Web~\cite{ramrakhya2022habitat}
& \cmark & 14.6* & 4.94* & -   & -    & -    & -    \\
ZSON~\cite{majumdar2022zson}
& \cmark & 15.3 & 4.8  & 25.5 & 12.6 & -    & -    \\
Navid~\cite{zhang2024navidvideobasedvlmplans}
& \cmark & -    & -    & 32.5 & 21.6 & -    & -    \\
PixNav~\cite{cai2024bridging}
& \cmark & -    & -    & 37.9 & 20.5 & -    & -    \\
ESC~\cite{zhou2023escexplorationsoftcommonsense}
& \cmark & 28.7 & 14.2 & 39.2 & 22.3 & -    & -    \\
PONI~\cite{ramakrishnan2022poni}
& \cmark & 31.8 & 12.1 & -    & -    & -    & -    \\
SemExp~\cite{chaplot2020object}
& \cmark & 36.0 & 14.4 & -    & -    & -    & -    \\
CompassNav~\cite{li2025compassnav}
& \cmark & 42.0 & 17.5 & 56.6 & 27.6 & -    & -    \\
Hydra-Nav-SFT~\cite{wang2026hydra}
& \cmark & 49.0 & 16.5 & -   & - & 72.9 & 27.7 \\

\midrule

UniGoal~\cite{yin2025unigoal}
& \cmark & 41.0 & 16.4 & -    & -    & 54.5 & 25.1 \\
Uni-Navid~\cite{zhang2024uni}
& \cmark & -    & -    & -    & -    & 73.7 & 37.1 \\
OmniNav~\cite{xue2025omninav}
& \cmark & -    & -    & -    & -    & 56.1 & 30.0 \\

\midrule

\rowcolor{gray!15}
\textbf{\ourmethod (Ours)}
& \cmark & \textbf{53.8} & \textbf{23.1}
& \textbf{70.5} & \textbf{34.6}
& \textbf{82.2} & \underline{38.5} \\

\bottomrule
\end{tabular}
\end{adjustbox}
\end{table*}

Table~\ref{tab:train_results} compares \ourmethod with representative training-free VLM systems, learning-based ObjectNav policies, and generic navigation models.
Among learning-based methods, \ourmethod achieves the best SR and SPL across all evaluated benchmarks.
Although trained only on MP3D demonstrations, \ourmethod transfers strongly to both HM3D-v1 and HM3D-v2.
Compared with existing learning-based policies, the substantial SPL gains suggest that learning from human demonstrations helps the agent adopt more efficient search behaviors.
For Hydra-Nav~\citep{wang2026hydra}, we report its supervised SFT configuration without iterative refinement, which is orthogonal to our contribution.
\ourmethod also attains the best SR among all compared methods on all three benchmarks, including training-free systems powered by larger API-hosted models; on HM3D-v2 its SPL (38.5) is second only to STRIVE (38.7) while achieving 2.6 points higher SR.
Unlike these API-based systems, the trained 2B model runs fully on-board with low inference latency (Sec.~\ref{sec:exp-realworld}).

\subsection{Ablation Study}
\label{sec:exp-ablation}

\paragraph{Decision interface.}
Table~\ref{tab:paradigm_lower_bound} examines how the decision interface affects performance on HM3D-v2.
The egocentric baseline predicts low-level action tokens from a uniformly sampled 5-frame RGB history.
The random baseline uses the same candidate set as \ourmethod, but randomly selects among frontier candidates until a verified target candidate is available.
The text-coordinate baseline predicts waypoint coordinates as free-form text.
These comparisons separate the benefit of BEV spatial grounding from the benefit of constrained candidate selection.
Random BEV candidate selection improves over egocentric action prediction, showing the \textbf{interface premium} of frontier and target candidates selection.
The reserved candidate-ID interface performs best, indicating that ranking discrete candidate options is more reliable than generating free-form waypoint coordinates.
Since the predicted viewpoints lie within previously explored regions, local path generation is straightforward; replacing the shortest-path follower with an FMM-based local planner changes SR by only 3\% over 500 episodes, indicating that the advantage of our method stems primarily from the learned high-level policy rather than privileged path following.

\paragraph{Policy components and objective.}

Table~\ref{tab:unified_ablation} ablates the main components of the spatial-visual candidate policy and the training objective on HM3D-v2.
Removing candidate visual memory causes the largest drop among deployable component variants, suggesting that egocentric evidence around each candidate is essential for assessing target relevance.
As a diagnostic upper-bound variant, replacing visual memory with oracle semantic labels further improves SR/SPL, indicating that more accurate semantic grounding could further improve performance.
Agent-centric geometry also plays an important role by binding candidate-level visual evidence to locations in the BEV decision space.
Replacing reserved candidate-ID tokens with text indices degrades performance, indicating that a constrained candidate vocabulary provides a more reliable selection interface.

For the objective, removing the angular soft target reduces both SR and SPL.
The full Intent-Aligned Objective improves over hard cross-entropy by smoothing frontier decisions according to directional proximity while preserving hard commitment for accepted target hypotheses.
It also reduces redundant revisits and increases exploration coverage relative to Hard CE Only, linking the objective to more directed search behavior.
Additional ablations and discussions are provided in Appendix~\ref{app:more-ablation}.

\paragraph{Necessity of training and BEV reliance.}
Table~\ref{tab:frozen_vlm} assesses whether the VLM backbone can natively perform candidate selection without our training.
Following the same execution policy as \ourmethod (committing to a selected frontier for at least six forward steps before re-querying), frozen InternVL3-2B repeatedly switches between nearby frontiers and oscillates in place, underperforming even random candidate selection, particularly in SPL.
Large-scale pretraining alone therefore does not expose candidate-selection competence, confirming the necessity of training.
To probe BEV reliance, we replace the BEV crop with a fully-unknown gray map at inference; SR/SPL drop from 82.2/38.5 to 56.8/20.5, close to the random-selection level.
With the component ablations in Table~\ref{tab:unified_ablation}, these results indicate that both the BEV map and candidate-wise grounding are crucial.

\begin{table}[t]
\centering
\small
\setlength{\tabcolsep}{5pt}
\caption{
\textbf{Decision-interface ablation on HM3D-v2.}
The results isolate the effect of BEV spatial-visual grounding and reserved candidate-ID selection.
}
\label{tab:paradigm_lower_bound}
\begin{tabular}{llcc}
\toprule
\textbf{Spatial interface}
& \textbf{Output format}
& \textbf{SR (\%)} $\uparrow$
& \textbf{SPL (\%)} $\uparrow$ \\
\midrule
Egocentric RGB history
& Action token
& 35.4 & 14.2 \\
BEV candidate space
& Random candidate
& 58.7 & 22.3 \\
BEV candidate space
& Text coordinate
& 68.4 & 26.9 \\
\midrule
\textbf{BEV candidate space}
& \textbf{Reserved candidate-ID token}
& \textbf{82.2} & \textbf{38.5} \\
\bottomrule
\end{tabular}
\end{table}
\begin{table}[t]
\centering
\small
\setlength{\tabcolsep}{4pt}
\caption{
\textbf{Component and objective ablation on HM3D-v2.}
We report SR/SPL and trajectory-level search behavior metrics for architecture and objective variants.
\emph{Visual} denotes the per-candidate visual signal: none, privileged ground-truth semantics, or RGB birth-view memory.
See Table~\ref{tab:search_behavior_mp3d} for a same-protocol MP3D comparison.
}
\label{tab:unified_ablation}
\resizebox{\linewidth}{!}{
\begin{tabular}{l cccc rr rrrr}
\toprule
& \multicolumn{4}{c}{\textit{Architecture}}
& \multicolumn{2}{c}{\textit{Performance}}
& \multicolumn{4}{c}{\textit{Search Behavior}} \\
\cmidrule(r){2-5}\cmidrule(lr){6-7}\cmidrule(l){8-11}
\textbf{Method}
& BEV & Visual & Geometry & Token
& SR\,(\%) $\uparrow$ & SPL\,(\%) $\uparrow$
& ECR $\uparrow$ & LSR $\uparrow$ & PMR $\uparrow$ & RRR $\downarrow$ \\
\midrule
\multicolumn{11}{l}{\small\textit{Interface baseline}} \\[0.3ex]
Ego-History Action Policy
  & \xmark & {--} & {--} & {--}
  & 35.4 & 14.2
  & 0.51 & 0.08 & 0.01 & 0.44 \\
\midrule
\multicolumn{11}{l}{\small\textit{Component ablation}} \\[0.3ex]
w/o Frontier Info
  & \cmark & None   & \cmark & \cmark
  & 70.2 & 29.4
  & 0.62 & 0.14 & 0.02 & 0.29 \\
Oracle Frontier Semantics
  & \cmark & GT Sem.& \cmark & \cmark
  & 84.6 & 40.2
  & 0.70 & 0.22
  & 0.04 & 0.16 \\
w/o Pairwise Geometry
  & \cmark & RGB    & \xmark & \cmark
  & 76.4 & 33.7
  & 0.66 & 0.17 & 0.03 & 0.24 \\
w/o Candidate ID Tokens
  & \cmark & RGB    & \cmark & \xmark
  & 78.6 & 35.1
  & 0.68 & 0.20 & 0.04 & 0.22 \\
\midrule
\multicolumn{11}{l}{\small\textit{Objective ablation}} \\[0.3ex]
Hard CE Only
  & \cmark & RGB    & \cmark & \cmark
  & 79.6 & 36.2
  & 0.70 & 0.19 & 0.03 & 0.20 \\
\textbf{\ourmethod}
  & \cmark & RGB    & \cmark & \cmark
  & 82.2 & 38.5
  & 0.71 & 0.21 & 0.05 & 0.18 \\
\midrule
\multicolumn{11}{l}{\small\textit{Human reference (MP3D, 300 eps.)}} \\[0.3ex]
Human Demo
  & {--} & {--} & {--} & {--}
  & \textit{92.7} & \textit{42.5}
  & \textit{0.60} & \textit{0.13} & \textit{0.07} & \textit{0.10} \\
\bottomrule
\end{tabular}
}
\end{table}

\begin{table}[t]
\centering
\small
\setlength{\tabcolsep}{4pt}
\caption{
\textbf{Frozen-backbone and BEV-masked ablations on HM3D-v2.}
\ourmethod (trained), BEV masked replaces only the BEV crop with a fully-unknown gray map at inference.
}
\label{tab:frozen_vlm}
\begin{tabular}{lcc}
\toprule
\textbf{Method} & \textbf{SR (\%)} $\uparrow$ & \textbf{SPL (\%)} $\uparrow$ \\
\midrule
Random candidate selection & 58.7 & 22.3 \\
Frozen InternVL3-2B: BEV + RGB & 55.8 & 14.2 \\
Frozen InternVL3-2B: BEV + RGB + geometry & 61.3 & 16.8 \\
\ourmethod (trained), BEV masked & 56.8 & 20.5 \\
\midrule
\textbf{\ourmethod (trained)} & \textbf{82.2} & \textbf{38.5} \\
\bottomrule
\end{tabular}
\end{table}

\subsection{Search Behavior Analysis}
\label{sec:exp-behavior}

Table~\ref{tab:unified_ablation} also reveals clear differences in trajectory-level search behavior.
Compared with the egocentric action baseline, all BEV-based variants substantially increase ECR, LSR, PMR and reduce RRR, indicating that the candidate-level BEV interface encourages broader spatial coverage and more effective search while mitigating locally repetitive behavior.
Within BEV-based variants, stronger spatial-visual grounding leads to more effective search behavior.
Removing visual memory reduces ECR and increases RRR, showing that candidate-level visual evidence helps the policy distinguish promising directions.
Oracle frontier semantics further improve SR/SPL and yield the lowest RRR among model variants, suggesting that accurate semantic grounding can make exploration more decisive.
Removing pairwise geometry or replacing reserved candidate-ID tokens also degrades both performance and behavior metrics, indicating that effective search requires not only visual evidence, but also a reliable spatial and decision interface for selecting among candidates.
The objective ablation further suggests that Intent-Aligned Objective promotes broader exploration, more active local scanning, and more purposeful short probing motions.

\begin{figure}[t]
\centering
\includegraphics[width=0.9\linewidth]{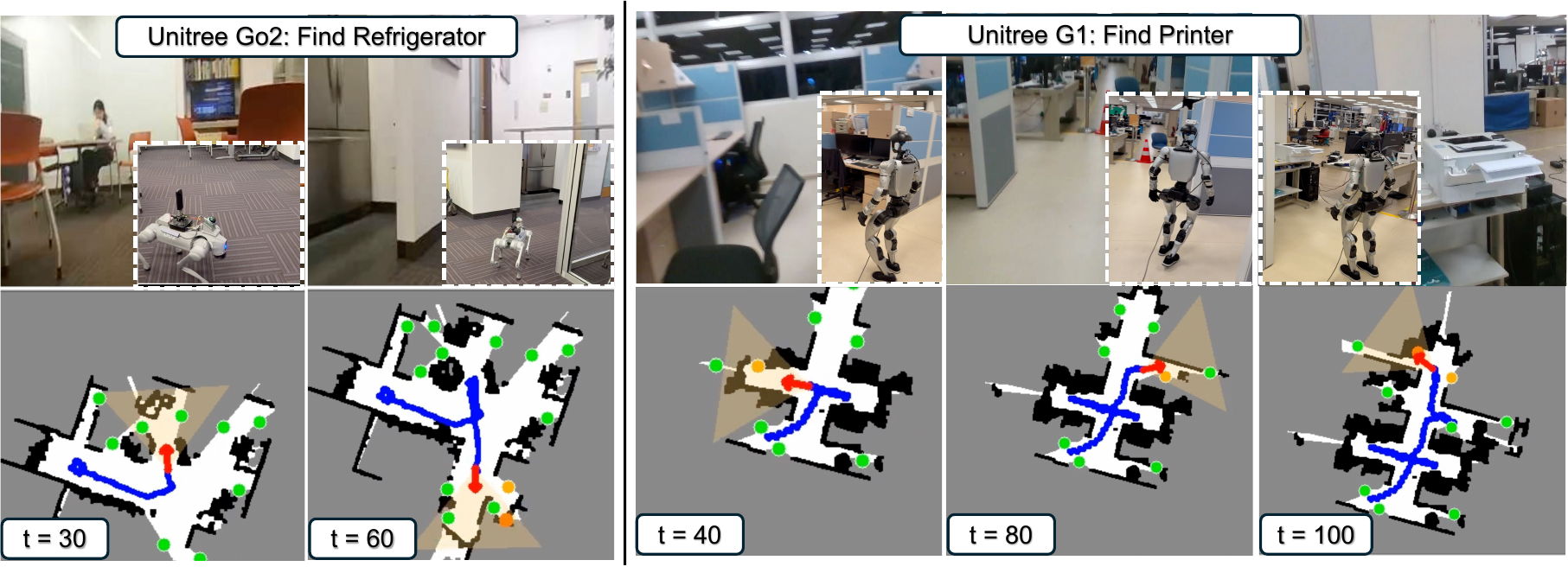}
\caption{
\textbf{Qualitative search behavior examples.}
\ourmethod expands unexplored regions through candidate waypoints, scans locally around informative viewpoints, and avoids redundant revisits.
}
\label{fig:case_study}
\end{figure}

We also evaluated 300 MP3D human demonstration episodes on the proposed metrics.
Humans achieve the lowest RRR and the highest PMR, reflecting decisive search with purposeful probing and few redundant revisits.
However, their ECR is lower than several learned policies because the demonstrations are success-only and typically terminate once the target is found.
Because this human reference is collected on MP3D, we further evaluate the ego-history action baseline and \ourmethod on MP3D under the same protocol (Table~\ref{tab:search_behavior_mp3d}).
The trends are consistent with HM3D-v2: stronger spatial-visual grounding yields broader coverage, more active scanning and probing, and substantially fewer redundant revisits.
Fig.~\ref{fig:case_study} qualitatively illustrates these behavioral patterns. Formal metric definitions and additional analysis are provided in Appendix~\ref{app:behavior-analysis}.

\begin{table}[t]
\centering
\small
\setlength{\tabcolsep}{4pt}
\caption{
\textbf{Search behavior analysis on MP3D-Val under the same evaluation protocol.}
The ego-history action baseline and \ourmethod are re-evaluated on MP3D; the human reference uses 300 sampled MP3D demonstration episodes.
}
\label{tab:search_behavior_mp3d}
\begin{tabular}{lcccc}
\toprule
\textbf{Method} & ECR $\uparrow$ & LSR $\uparrow$ & PMR $\uparrow$ & RRR $\downarrow$ \\
\midrule
Ego-History Action Policy & 0.44 & 0.07 & 0.01 & 0.48 \\
\textbf{\ourmethod} & \textbf{0.68} & \textbf{0.19} & 0.05 & 0.19 \\
Human Demo (MP3D, 300 eps.) & 0.60 & 0.13 & \textbf{0.07} & \textbf{0.10} \\
\bottomrule
\end{tabular}
\end{table}

\begin{table}[!htbp]
\centering
\small
\setlength{\tabcolsep}{5pt}
\caption{
\textbf{Real-world deployment statistics.}
The same MP3D-trained checkpoint is deployed on all three platforms without robot-specific fine-tuning.
}
\label{tab:real_world_summary}
\begin{tabular}{lc}
\toprule
\textbf{Statistic} & \textbf{Value} \\
\midrule
Total trials (wheeled / Go2 / G1) & 55 (28 / 17 / 10) \\
Overall SR (\%) & 52.7 \\
SR (\%) per platform (wheeled / Go2 / G1) & 64.3 / 41.2 / 40.0 \\
\bottomrule
\end{tabular}
\end{table}

\subsection{Real-World Deployment}
\label{sec:exp-realworld}

We deploy the same MP3D-trained checkpoint to three physical embodiments: a wheeled robot, a Unitree Go2 quadruped, and a Unitree G1 humanoid, without additional VLM fine-tuning.
All deployments run the VLM policy on an RTX 4090 laptop and use the same candidate-level BEV waypoint interface, with platform-specific sensing calibration and local control.
The BEV map is constructed online from onboard sensing---RGB-D in simulation and LiDAR with SLAM on the robots---so no pre-built environment map is required (Appendix~\ref{app:physical-deployment}).
Unlike training-free pipelines that often rely on multi-stage prompting or repeated VLM queries, \ourmethod performs low-frequency candidate-ID prediction over a compact waypoint set.
At runtime, the next policy query is triggered as the robot approaches the current waypoint, allowing VLM inference to overlap with waypoint execution rather than pausing the robot for each query.
Each high-level policy inference takes approximately $0.8$--$1.4\,\mathrm{s}$ depending on the candidate count (Appendix~\ref{app:cost}), enabling real-time deployment.
We test on 8 target object categories across 10 different real-world scenes.

Table~\ref{tab:real_world_summary} reports quantitative statistics over all deployment trials.
Evaluated scenarios typically span multiple rooms, with start-to-goal distances of 10--30\,m.
Most failures trace to deployment-level perception, localization, and control issues---detector misses, SLAM pose drift, and planning failures in tight passages---rather than high-level search decisions.
More details are provided in Appendix~\ref{app:physical-deployment}.


\section{Conclusion}
\label{sec:conclusion}

We presented \ourmethod, a BEV-grounded spatial-visual imitation framework that learns candidate-level ObjectNav decisions from human demonstrations.
Frontier-based Human-Intent Labeling converts low-level human trajectories into waypoint-level proxy labels, and an Intent-Aligned Objective encourages directionally consistent exploration and decisive target commitment.
Trained only on MP3D demonstrations, \ourmethod achieves state-of-the-art performance among learning-based methods across MP3D, HM3D-v1, and HM3D-v2, and transfers zero-shot to wheeled, quadruped, and humanoid robots through the same candidate-level BEV waypoint interface.

\paragraph{Limitations.}
\ourmethod assumes a mostly static environment, so dynamic changes may make stored map regions and birth-view memories stale; execution failures and detours are also not yet fed back into the VLM policy, as waypoint selection is decoupled from platform-specific local planning.
Future work should incorporate temporal memory updates for dynamic scenes and tighter feedback between high-level selection and low-level execution.


\clearpage
\acknowledgments{
We thank the CoRL 2026 Area Chair and reviewers for their constructive feedback, which helped improve this paper.
Chen Lv is partially by Ministry of Education (MOE) of Singapore under NTUitive Gap Fund (NGF-2025-18-033). 
Wei-Yun Yau is partially supported by A*STAR under the National Robotics Programme grant no.~H25-MNR3495.
}


\bibliography{ref}

\clearpage

\appendix

\begin{center}
    {\LARGE \textbf{Appendix}}
\end{center}
\vspace{1em}

\section{BEV Map Construction from RGB-D Observations}
\label{app:bev-map}

This section describes the BEV mapping procedure used to construct the spatial input $M_t$ and the frontier set in Sec.~\ref{sec:method-candidate-policy}. The mapping module maintains a global occupancy map with resolution $r=0.05\,\mathrm{m}$ per cell. The global map covers $67.2\,\mathrm{m}\times 67.2\,\mathrm{m}$, corresponding to a $1344\times 1344$ grid, and the VLM receives a $448\times 448$ egocentric crop covering $22.4\,\mathrm{m}\times 22.4\,\mathrm{m}$. At each timestep, the agent receives an RGB-D observation and the camera and agent pose from Habitat. We filter the raw depth map to the valid interval $[0.51, 4.9]\,\mathrm{m}$ and fill small holes caused by thin structures or depth discontinuities. Valid depth pixels are then back-projected into a camera-frame point cloud using the camera intrinsic matrix and transformed into the world frame using the current camera extrinsics.

Let $P_t^w=\{(x_j,y_j,z_j)\}$ denote the world-frame point cloud after depth filtering and camera-to-world transformation. We project a point to grid coordinates by
\begin{equation}
u_j = \left\lfloor \frac{x_j + m_0}{r} \right\rfloor,\qquad
v_j = \left\lfloor \frac{y_j + m_0}{r} \right\rfloor,
\label{eq:bev-grid-projection}
\end{equation}
where $m_0$ is the map center offset and $(v_j,u_j)$ are row-column grid indices. We set the floor height to $h_0=0.0\,\mathrm{m}$. Points near the floor, $z_j < h_0 + 0.2\,\mathrm{m}$, support the explored/free-space estimate. Points in the obstacle band, $h_0+0.2\,\mathrm{m} < z_j < 1.5\,\mathrm{m}$, are projected to the floor plane and support the occupied-cell estimate. Points below $h_0-0.2\,\mathrm{m}$ and points above $1.5\,\mathrm{m}$ are filtered out when constructing obstacle evidence, which removes floor noise and high ceiling structure. The map therefore maintains two complementary binary layers: an occupancy layer $O_t$ for non-traversable cells and an explored layer $E_t$ for cells observed by the sensor. This distinction is important because an unoccupied but unobserved cell should not be treated as known free space.

\paragraph{Depth-column completion.} First-person depth images often contain vertical holes around thin structures (e.g., table legs), where background depth shows through and causes false free space in the BEV projection. Before back-projection, we scan each column from the bottom row to the image midpoint---the upper half is skipped as it typically captures ceiling or sky. At each pixel $j$, a discontinuity is flagged when the pixel above is valid foreground ($d_{j-1} > 0.58\,\mathrm{m}$) but much closer than the current pixel ($d_{j-1} < d_j - 0.08\,\mathrm{m}$). We then find the nearest row below $j$ where depth returns to $d_{j-1}$ and overwrite all farther pixels in between with $d_{j-1}$.

\paragraph{Near-field floor completion.} The first-person camera leaves a near-field blind region on the floor. With $640\times480$ resolution, $79^\circ$ horizontal field of view, and $0.88\,\mathrm{m}$ camera height, the vertical field of view is $\phi_v \approx 63.45^\circ$ and the lowest camera ray intersects the floor at
\begin{equation}
d_{\mathrm{blind}} = \frac{0.88}{\tan(\phi_v/2)} \approx 1.42\,\mathrm{m},
\end{equation}
leaving roughly $28.5$ BEV cells directly in front of the agent unobserved. If untreated, this blind region creates artificial unexplored holes and spurious frontiers around the agent. We fill it conservatively: within the horizontal field of view, we discretize bearing into $3^\circ$ bins, find the nearest observed floor or obstacle point in each bin (up to $2.0\,\mathrm{m}$ from the agent), and interpolate along the segment from the agent to that point, marking the intervening BEV cells as explored. This only affects the explored layer; obstacle cells are still determined by the obstacle-height point set above.

\begin{algorithm}[t]
\caption{BEV map update from an RGB-D observation}
\label{alg:bev-map-update}
\begin{algorithmic}[1]
\Require RGB image $I_t$, depth image $D_t$, camera pose $T^w_{c,t}$, agent pose $T^w_{a,t}$, previous map $M_{t-1}$
\Ensure Updated global map $M_t$ and local BEV crop
\State Filter invalid depth values outside $[0.51,4.9]\,\mathrm{m}$ and fill small depth holes.
\State Back-project valid depth pixels into a camera-frame point cloud $P^c_t$.
\State Transform $P^c_t$ into the world frame: $P^w_t \leftarrow T^w_{c,t} P^c_t$.
\State Project $P^w_t$ to grid coordinates using the map resolution.
\State Mark cells supported by floor-level points ($z<h_0+0.2\,\mathrm{m}$) as explored.
\State Mark cells supported by obstacle-band points ($h_0+0.2<z<1.5\,\mathrm{m}$) as occupied and explored.
\State For nearby observed boundaries within $2.0\,\mathrm{m}$, interpolate short rays from the agent to fill local explored free space.
\State Project the current agent pose to the grid and update the agent-footprint and trajectory layers.
\State Crop the global map around the current agent pose to obtain the local BEV input.
\end{algorithmic}
\end{algorithm}

\paragraph{Map channels.} The local BEV crop contains four geometric channels used throughout the paper: occupancy, explored area, current agent footprint, and trajectory history. The occupancy channel indicates cells that are estimated to be blocked by obstacles. The explored channel indicates cells that have been observed by the sensor, regardless of whether they are free or occupied. The agent-footprint channel marks where the agent is at the current timestep, while the trajectory channel accumulates previously visited footprints. For training-time analysis and some oracle variants, semantic evidence can also be accumulated in additional per-category channels by projecting semantic predictions or labels into the same grid; the main learned policy described in Sec.~\ref{sec:method-candidate-policy} uses the rendered BEV crop together with waypoint-specific RGB memory.

\paragraph{Frontier extraction.} Frontiers are extracted from the occupancy and explored layers of the BEV map. We first identify boundary cells between explored free space and unexplored regions, then apply simple morphological filtering and connected-component filtering to remove small noisy components and frontiers too close to the map border. The remaining frontier components are represented by their centers and converted into local BEV coordinates. These centers form the frontier portion of the waypoint option set $\mathcal{C}_t$.

\section{Waypoint Supervision Strategies}
\label{app:frontier-collapse}

Human demonstrations are recorded as low-level action sequences rather than explicit waypoint decisions, so we require a label-recovery procedure to convert each replayed step into a candidate-level supervision label $y_t$.
This section compares three recovery strategies over the same candidate set $\mathcal{C}_t$.
For all strategies, if a verified target candidate is present in $\mathcal{C}_t$, its index is used directly as the label.

\subsection{Strategy 0: Pose-Heading Matching}

This local heuristic selects the frontier most aligned with the agent's instantaneous heading, without looking into the future trajectory.
Let $\mathbf{p}_a$ and $\mathbf{h}_a$ denote the agent position and unit heading vector in BEV pixel coordinates.
For each frontier candidate $x_i$, we compute the angle between $\mathbf{h}_a$ and the normalized direction $(x_i-\mathbf{p}_a)/\|x_i-\mathbf{p}_a\|_2$, and select the candidate with the smallest angular deviation:
\begin{equation}
y_t = \arg\min_i\; \arccos\!\left(\mathrm{clip}\!\left(\mathbf{h}_a^\top \frac{x_i-\mathbf{p}_a}{\|x_i-\mathbf{p}_a\|_2},\,-1,\,1\right)\right).
\end{equation}
This is simple but noisy when the demonstrator is turning in place, scanning the scene, or temporarily facing away from the intended frontier.

\subsection{Strategy 1: Future-Trajectory Matching}

This strategy compares each frontier candidate with the remaining demonstration path $\tau_{t:T}$ (from step $t$ to the episode end $T$).
For each frontier $x_i$, we compute a shortest path from $\mathbf{p}_a$ to $x_i$ via the Habitat-Sim pathfinder, then measure a one-sided Chamfer distance from the future trajectory to this path:
\begin{equation}
y_t =
\arg\min_i
\frac{1}{|\tau_{t:T}|}\sum_{p \in \tau_{t:T}} \min_{q \in \mathrm{Path}(\mathbf{p}_a,x_i)} \|p - q\|_2 .
\end{equation}
This is more robust to instantaneous orientation noise than pose-heading matching, but remains ambiguous when multiple frontiers lie along similar partial trajectories or when the demonstrator performs short probing motions before committing to a direction.

\begin{table}[t]
\centering
\small
\setlength{\tabcolsep}{5.0pt}
\caption{
\textbf{Human-intent label recovery ablation on MP3D-Val.}
We compare strategies for converting low-level human demonstrations into candidate-level waypoint labels, holding the candidate set, model architecture, inputs, and objective fixed.
}
\label{tab:ablation_label}
\begin{tabular}{lcc}
\toprule
\textbf{Label recovery strategy}
& \textbf{SR (\%) $\uparrow$}
& \textbf{SPL (\%) $\uparrow$} \\
\midrule
Strategy 0: Pose-heading matching              & 38.9 & 17.1 \\
Strategy 1: Future-trajectory matching         & 42.0 & 20.6 \\
Strategy 2: \textbf{Frontier-collapse (Ours)}  & \textbf{53.8} & \textbf{23.1} \\
\bottomrule
\end{tabular}
\end{table}

\subsection{Strategy 2: Frontier-Collapse Supervision}

The frontier-collapse strategy (Sec.~\ref{sec:method-candidate-policy}) selects the frontier whose disappearance from the future frontier set is most pronounced. Algorithm~\ref{alg:frontier-collapse} details the procedure: we advance the replay to a pivot step $t^+$ (at least $K{=}6$ additional \textsc{move\_forward} actions, extended if the frontier set is still nearly unchanged), then pick the current non-target frontier with the largest geodesic nearest-neighbor distance to $\mathcal{F}_{t^+}$ in world coordinates.

\begin{algorithm}[h]
\caption{Frontier-collapse label recovery}
\label{alg:frontier-collapse}
\begin{algorithmic}[1]
\Require Demonstration replay $\tau$, current step $t$, current waypoint set $\mathcal{C}_t$, future frontier sets $\{\mathcal{F}_{t'}\}_{t'>t}$
\Ensure Recovered waypoint label $y_t$
\If{a valid target option exists in $\mathcal{C}_t$}
    \State \Return index of the target option
\EndIf
\State Initialize $t^+ \leftarrow t$ and forward-action counter $m\leftarrow 0$
\While{$m < K$ and $t^+$ is not the final replay step}
    \State $t^+ \leftarrow t^+ + 1$
    \If{the action at step $t^+-1$ is \textsc{move\_forward}}
        \State $m \leftarrow m+1$
    \EndIf
\EndWhile
\While{$t^+$ is not the final replay step and the future frontier set has not changed sufficiently}
    \State $t^+ \leftarrow t^+ + 1$
    \If{the action at step $t^+-1$ is \textsc{move\_forward}}
        \State $m \leftarrow m+1$
    \EndIf
\EndWhile
\For{each non-target waypoint option $x_i \in \mathcal{C}_t$}
    \State Convert $x_i$ from local BEV coordinates to world coordinates.
    \State Convert each future frontier $f\in\mathcal{F}_{t^+}$ to world coordinates.
    \State $d_i \leftarrow \min_{f\in\mathcal{F}_{t^+}} \mathrm{geo}(x_i,f)$ using the Habitat-Sim pathfinder.
\EndFor
\If{all frontiers are unreachable or $\max_i d_i$ is below a collapse threshold}
    \State Fall back to future-trajectory matching.
\Else
    \State $y_t \leftarrow \arg\max_i d_i$
\EndIf
\State \Return $y_t$
\end{algorithmic}
\end{algorithm}

Two safeguards handle degenerate cases: (i)~a verified target option, if present, is used directly; (ii)~if all frontiers are unreachable or the maximum collapse distance is below threshold, we fall back to Strategy~1. These cases are rare.

\paragraph{Ablation.}
Table~\ref{tab:ablation_label} compares the three strategies on MP3D-Val with all other factors held constant.
Pose-heading matching suffers from instantaneous orientation noise; future-trajectory matching improves upon it but remains path-geometry-based; frontier-collapse performs best, suggesting that frontier evolution provides the most reliable signal for recovering exploration intent.

\section{Target Proposal and Verification}
\label{app:target-verification}

The target option is treated as a verified waypoint hypothesis rather than a raw detection, because detections may be transient, class-confusable, or geometrically unsuitable for navigation.
Across training replay, simulation evaluation, and physical deployment, the policy receives the same interface: a target candidate is appended to the frontier set only after it is semantically valid, temporally stable, geometrically localizable, and near a navigable position.

During training replay, we use simulator semantic annotations to construct target hypotheses, isolating high-level search-intent learning from detector noise.
During evaluation and deployment, targets are instead produced by online object detectors.
Once accepted, a target is continuously verified; if it should be visible but is not re-detected for multiple frames, it is removed and the policy returns to frontier exploration.

\paragraph{Target proposal source.}
Training replay uses simulator semantic masks to check whether goal-category pixels are visible.
Evaluation uses online detectors: an open-vocabulary detector-segmenter in simulation, and a robot-side YOLOE module (Appendix~\ref{app:physical-deployment}) in physical deployment.
ObjectNav categories are mapped to detector aliases (e.g., \verb|sofa|$\to$\verb|couch|, \verb|plant|$\to$\verb|potted_plant|).
For visually confusable goals, the detector is also prompted with nearby categories (e.g., \verb|chair|, \verb|bench|, \verb|stool| for chair-like goals).

\paragraph{Semantic and temporal verification.}
A proposal is positive only when its class matches the target aliases.
We maintain a binary detection buffer over the last three frames and require at least two positives before accepting.
After acceptance, the target mask is converted to a target-specific depth image and back-projected into a 3D point cloud using the same camera intrinsics and extrinsics as the BEV mapper.

\paragraph{3D localization and navigable waypoint selection.}
The target point cloud is voxel-downsampled and reduced to its largest connected component; if none remains, the hypothesis is discarded.
We compute two 3D summaries: the component median $A$, and the median of the $K{=}\min(5,N)$ nearest-to-agent points $B$.
The raw target waypoint is
\begin{equation}
\mathbf{p}_{\mathrm{tar}} = 0.7B + 0.3A,
\end{equation}
biased toward the visible side of the object.
We reject if $\mathbf{p}_{\mathrm{tar}}$ is more than $0.5\,\mathrm{m}$ below the agent.
We then search for a nearby navigable point on the same navigation island: first retreating toward the agent in $0.1\,\mathrm{m}$ steps, then probing radial offsets if retreat fails.
The maximum allowed shift is $1.0\,\mathrm{m}$ by default, $1.5\,\mathrm{m}$ for toilets, and $2.0\,\mathrm{m}$ for potted plants and TV monitors.
Accepted waypoints are projected to the BEV grid and appended to the candidate set if within the local crop.

\paragraph{Persistence check during approach.}
While following a target, we maintain a missed-detection counter that is incremented only when the target is expected to be observable (inside the camera frustum and not occluded by obstacles).
When the counter reaches $N_{\mathrm{miss}}$, the hypothesis is invalidated and removed from the waypoint set.
This prevents commitment to a stale false positive while avoiding unnecessary deletion when the target is temporarily out of view or occluded.

\begin{algorithm}[t]
\caption{Target proposal and verification}
\label{alg:target-verification}
\begin{algorithmic}[1]
\Require RGB-D observation $(I_t,D_t)$, current map state, target category $g$, recent detection buffer $\mathcal{B}$, active target hypothesis $H$
\Ensure Updated target hypothesis $(\texttt{target\_found}, x_{\mathrm{tar}})$
\State Obtain target proposals from the phase-specific source: simulator semantic masks for training replay, or online detector masks for evaluation/deployment.
\State Map $g$ to detector aliases when using an online detector.
\State Filter proposals by class/alias match, confidence, mask area, and class-specific priors.
\State Set $b_t\leftarrow 1$ if at least one proposal matches the target category; otherwise set $b_t\leftarrow 0$.
\If{$H$ is active}
\If{$H$ is inside the camera-facing sector and is not occluded}
\If{$b_t=0$}
\State Increase the missed-detection counter for $H$.
\Else
\State Reset the missed-detection counter for $H$.
\EndIf
\EndIf
\If{the missed-detection counter reaches $N_{\mathrm{miss}}$}
\State Clear $H$ and remove the target from the waypoint set.
\EndIf
\EndIf
\State Append $b_t$ to the length-$3$ detection buffer $\mathcal{B}$.
\If{$\sum_{b\in\mathcal{B}} b < 2$}
\State \Return no accepted target hypothesis.
\EndIf
\State Back-project the verified target mask and depth values to a world-frame point cloud.
\State Voxel-downsample the point cloud and keep its largest connected component.
\If{the component is empty or too small}
\State \Return no accepted target hypothesis.
\EndIf
\State Compute $A$ as the component median and $B$ as the median of the $K=\min(5,N)$ nearest component points to the agent.
\State $\mathbf{p}_{\mathrm{tar}}\leftarrow 0.7B+0.3A$.
\If{$\mathbf{p}_{\mathrm{tar}}$ is more than $0.5\,\mathrm{m}$ below the agent height}
\State \Return no accepted target hypothesis.
\EndIf
\State Move $\mathbf{p}_{\mathrm{tar}}$ toward the agent in $0.1\,\mathrm{m}$ steps until it is navigable and on the same navigation island.
\If{the shifted point is too far from the visible target}
\State Probe radial offsets around the target in $0.1\,\mathrm{m}$ rings and keep the first navigable point on the same island.
\EndIf
\If{the final navigable point is still farther than the category-specific distance threshold}
\State \Return no accepted target hypothesis.
\EndIf
\State Project the accepted 3D point to the BEV grid and append it as the target waypoint candidate if it lies in the local crop.
\end{algorithmic}
\end{algorithm}

\paragraph{Interaction with waypoint selection.}
Once accepted, the target uses the same interface as frontier waypoints: rendered on the BEV map, assigned a local coordinate, and inserted into the candidate list.
If later invalidated, it is removed and the policy continues with frontier-only candidates.
During execution, a shortest-path follower moves toward the selected waypoint.
If the waypoint is a frontier and the follower reaches it, the system requests a new VLM decision rather than stopping.
A stop is issued only when the followed waypoint is the verified target, preventing frontier arrival from being mistaken for task completion.

\section{Physical-Robot Deployment Adaptations}
\label{app:physical-deployment}

The learned waypoint policy is kept unchanged in physical deployment; only the simulator-specific perception and control interfaces are replaced. A ROS~2 bridge converts registered LiDAR scans, SLAM odometry, and egocentric camera images into the same BEV and candidate representation used during training, then publishes the selected waypoint to the robot's local planner.

\paragraph{LiDAR-based BEV construction.} The robot subscribes to SLAM-registered LiDAR scans and odometry rather than dense depth images. The global BEV map preserves the training geometry ($67.2\,\mathrm{m}{\times}67.2\,\mathrm{m}$, $0.05\,\mathrm{m}$ resolution, $448{\times}448$ crop, four channels). To reduce distribution shift, LiDAR returns are filtered to the $79^\circ$ camera-facing sector. Obstacle evidence uses a robot-relative height band $[-0.4, 1.2]\,\mathrm{m}$; explored cells are marked by ray casting that stops at occupied cells to prevent leaking through walls. Near-body returns within $0.1\,\mathrm{m}$ and isolated speckles (voxel support filter at $0.10\,\mathrm{m}$, $\geq 2$ points per voxel) are removed.

\paragraph{Frontier stabilization for real scans.} Frontier extraction follows the same explored--unexplored boundary definition as in simulation, with additional filters for noisy LiDAR maps. The extractor operates on the global map to avoid local-crop flicker, keeps only the robot-connected explored component, dilates obstacles before frontier masking to suppress corner artifacts, removes tiny components and candidates near the crop border or within $0.7\,\mathrm{m}$ of the robot, merges close duplicates ($1.0\,\mathrm{m}$ threshold), and passes at most 32 candidates to the VLM.

\paragraph{Camera projection and visual memory.} The bridge projects the panoramic camera stream into a $640{\times}480$ pinhole view with $79^\circ$ horizontal field of view, accounting for robot heading, camera yaw offset, and camera-to-body translation. For each frontier, a birth RGB image is stored keyed by quantized world position ($0.5\,\mathrm{m}$ resolution) and provided to the ego ViT in candidate order. An accepted target's first observed RGB frame is appended as the final candidate image, matching the simulation convention.

\paragraph{External target detector.} Detection runs in a separate ROS node using a YOLOE segmentation model (TensorRT engine), publishing JSON detections with label, confidence, bounding box, and optional mask. The bridge filters detections by goal category and confidence (detector $\geq 0.3$, acceptance $\geq 0.4$). Detection timestamps are aligned with SLAM poses via short pose-history interpolation, which is critical because even small delays can place the target bearing on the wrong side of the BEV map.

\paragraph{Target localization without metric depth.} The physical detector provides no depth map, so we localize the target by ray casting along the bounding-box center bearing in the LiDAR BEV obstacle map. The ray origin is offset to the camera optical center using the calibrated translation; the ray stops at the first occupied cell. If no hit is found, the target is not locked. Otherwise the hit point is projected into the local BEV crop and appended after the frontier candidates, giving the VLM the same target interface as in simulation.

\paragraph{ROS waypoint execution and replanning.} The bridge publishes the selected waypoint as a \texttt{PointStamped} on \texttt{/way\_point}; the robot's local planner handles collision avoidance and velocity control. As in simulation (Appendix~\ref{app:target-verification}), approaching within $0.5\,\mathrm{m}$ of a frontier waypoint triggers an immediate VLM re-query rather than a full stop, allowing inference to overlap with the final approach. Reaching a target waypoint publishes \texttt{/vlm\_target\_reached} and holds position. Fresh target detections can trigger an immediate VLM query, allowing the policy to switch from exploration to target approach as soon as a localizable target appears.

\section{Additional Objective Ablation}
\label{app:more-ablation}

The main paper compares the full Intent-Aligned Objective with hard cross-entropy.
Here, we further separate two design choices: whether angular soft supervision is used, and whether it is applied to target-visible steps.
Frontier candidates benefit from soft supervision because nearby directions are plausible alternatives, but smoothing toward nearby frontiers weakens commitment to a verified target.

\begin{table}[t]
    \centering
    \caption{
    \textbf{Objective ablation on HM3D-v2.}
    We compare hard supervision, angular soft supervision applied to all steps, and the proposed target-safe angular objective.
    Target Commit Rate measures how often the model selects the target candidate when it is available.
    }
    \label{tab:objective_ablation}
    \resizebox{\linewidth}{!}{
    \begin{tabular}{lccccc}
        \toprule
        Objective
        & Frontier Steps
        & Target-Visible Steps
        & SR (\%) $\uparrow$
        & SPL (\%) $\uparrow$
        & Target Commit Rate (\%) $\uparrow$ \\
        \midrule
        Hard CE Only
        & Hard
        & Hard
        & 79.6 & 36.2 & 91.0 \\
        Angular Soft CE Everywhere
        & Soft
        & Soft
        & 77.8 & 34.6 & 84.3 \\
        \textbf{Target-Safe Angular Soft CE}
        & Soft
        & Hard
        & \textbf{82.2} & \textbf{38.5} & \textbf{94.8} \\
        \bottomrule
    \end{tabular}
    }
\end{table}

Table~\ref{tab:objective_ablation} shows that applying angular soft supervision to all steps hurts both SR/SPL and target commitment.
Hard CE preserves commitment but ignores the directional structure of frontier alternatives.
The proposed target-safe angular objective combines both: angular soft supervision for frontier decisions and hard supervision for target candidates, yielding the best SR, SPL, and Target Commit Rate.

\section{Prompt Template}
\label{app:prompt}

This section documents the prompt template used to serialize the BEV map, candidate set, target goal, and candidate-associated egocentric RGB memories into the VLM input.
The template is fixed across training and evaluation; only the runtime contents change.
Ablation variants differ only in geometry injection form or output vocabulary.

\paragraph{Special tokens.}
\ourmethod adds dedicated tokens: \texttt{<image\_bev>} and \texttt{<image\_ego>} for the BEV and ego vision encoders; \texttt{<state>}/\texttt{<s>}/\texttt{<e\_s>} to delimit the agent-state block; \texttt{<candidates>}/\texttt{<cand>}/\texttt{<e\_cand>} to delimit the candidate list.
We further reserve $32$ candidate-ID output tokens \texttt{<id\_0>}--\texttt{<id\_31>}; the prediction in Eq.~\ref{eq:policy} is a restricted softmax over only the active ID tokens.

\paragraph{System / task instruction.}
The prompt opens with a fixed instruction:
\begin{quote}\small\ttfamily
Imagine you are an autonomous robot in an indoor habitat environment.\\
Inputs:\\
- BEV grid map <image\_bev> showing free space (white), occupied space (black), unexplored area (gray), frontier candidates (green dots), robot pose/heading (red arrow), past trajectory (blue line), and egocentric camera field of view (yellow cone). An orange dot may appear on the BEV map indicating the detected goal location.\\
- Each candidate is paired with an egocentric RGB memory image <image\_ego>, which provides local visual context around that candidate.\\
- Goal: search for and navigate to **\{goal\}**.
\end{quote}
\noindent The placeholder \texttt{\{goal\}} is replaced with the ObjectNav target category, such as \texttt{chair}, \texttt{tv\_monitor}, or \texttt{plant}.

\paragraph{State block.}
The agent pose is serialized as a single line in BEV-grid pixel coordinates:
\begin{quote}\small\ttfamily
<state> <s> pos=(r, c) yaw\_deg=$\theta$ <e\_s>
\end{quote}

\paragraph{Candidate block.}
Each waypoint is serialized as one \texttt{<cand>}~\dots~\texttt{<e\_cand>} block containing the ID token, type, BEV coordinate, view label, and RGB memory image:
\begin{quote}\small\ttfamily
<cand> id\_token=<id\_k> type=\{frontier$|$target\} pos=(r, c) \{Frontier view$|$Target view\} <image\_ego> <e\_cand>
\end{quote}
\noindent For frontier candidates, \texttt{<image\_ego>} is the RGB frame when the frontier was first discovered; for targets, it is the frame of the accepted hypothesis.
The relative-geometry feature $\mathbf{f}_i$ (Eq.~\ref{eq:rel-feat}) is injected at the embedding level through the pairwise spatial encoder, added to \texttt{<cand>} and, when dual text-position injection is enabled, also to \texttt{<e\_cand>}.
Frontier and target candidates share the same template, differing only in \texttt{type} and view label.

\paragraph{Output instruction.}
The prompt closes with:
\begin{quote}\small\ttfamily
Choose one candidate token. Output only one token in the form <id\_k>.
\end{quote}

\paragraph{Assembled example.}
A full prompt with three frontiers and one target:
\begin{quote}\small\ttfamily
Imagine you are an autonomous robot in an indoor habitat environment.\\
Inputs:\\
- BEV grid map <image\_bev> showing free space (white), occupied space (black), unexplored area (gray), frontier candidates (green dots), robot pose/heading (red arrow), past trajectory (blue line), and egocentric camera field of view (yellow cone). An orange dot may appear on the BEV map indicating the detected goal location.\\
- Each candidate is paired with an egocentric RGB memory image <image\_ego>, which provides local visual context around that candidate.\\
- Goal: search for and navigate to **tv\_monitor**.\\
\\
<state> <s> pos=(224, 224) yaw\_deg=90.0 <e\_s>\\
\\
<candidates>\\
<cand> id\_token=<id\_0> type=frontier pos=(180, 200) Frontier view <image\_ego> <e\_cand>\\
<cand> id\_token=<id\_1> type=frontier pos=(220, 280) Frontier view <image\_ego> <e\_cand>\\
<cand> id\_token=<id\_2> type=frontier pos=(150, 160) Frontier view <image\_ego> <e\_cand>\\
<cand> id\_token=<id\_3> type=target pos=(228, 232) Target view <image\_ego> <e\_cand>\\
</candidates>\\
\\
Choose one candidate token. Output only one token in the form <id\_k>.
\end{quote}
\noindent The expected output is \texttt{<id\_3>}, committing to the verified target.

\paragraph{Training vs.~inference.}
The prompt structure is identical in both phases.
Two details differ: (i)~at training time, candidate order is randomly permuted and the supervision target is remapped accordingly, preventing positional shortcuts; at inference, the natural order is used (frontiers first, target last). (ii)~The angular soft target (Eq.~\ref{eq:angular-target}) is applied only during training on frontier-selection steps; at inference, the model argmax-decodes a single \texttt{<id\_k>} token.
No few-shot or in-context examples are used.

\section{Search Behavior Analysis}
\label{app:behavior-analysis}

This appendix expands the search-behavior summary in Sec.~\ref{sec:exp-behavior}.
We define four trajectory-level diagnostic metrics computed offline from the agent state log, BEV exploration history, and the scene navmesh. These metrics are used only for analysis.

\paragraph{Metric definitions.}
\begin{enumerate}[leftmargin=*,itemsep=1pt]
\item \textbf{Exploration Coverage Ratio} (ECR): fraction of reachable free-space area (from the navmesh) observed before episode termination.
\item \textbf{Local Scanning Ratio} (LSR): fraction of length-$W{=}6$ windows where translation $<0.3\,\mathrm{m}$ and accumulated heading change $>60^\circ$.
\item \textbf{Probing Motion Ratio} (PMR): fraction of length-$W{=}15$ windows where the agent moves $>1.0\,\mathrm{m}$ from the window start but returns within $0.5\,\mathrm{m}$ by the window end.
\item \textbf{Redundant Revisit Ratio} (RRR): fraction of steps that both (i)~do not expand the explored area ($\Delta\mathrm{explored}\le 0$) and (ii)~bring the agent within $0.3\,\mathrm{m}$ of a previously visited position.
\end{enumerate}
Higher ECR/LSR/PMR indicate more active exploration; lower RRR indicates fewer unproductive revisits.

\paragraph{Self-comparison on HM3D-v2.}
Table~\ref{tab:unified_ablation} in the main paper reports these metrics; the main text summarizes the overall monotonic co-variation with SR/SPL.
Here we highlight notable \emph{deviations} from that trend.
Removing candidate-ID tokens raises LSR despite lower SR, suggesting that less constrained outputs induce extra local scanning before commitment.
Oracle Frontier Semantics achieves the highest SR ($84.6$) yet slightly lower ECR ($0.70$ vs.\ $0.71$), consistent with finding targets earlier and terminating sooner; its near-identical LSR/PMR to \ourmethod suggests the remaining SR gap is driven by semantic grounding rather than exploration pattern.

These deviations also illustrate why the metrics should not be interpreted as universally monotonic objectives.
Higher ECR is not always better if an agent succeeds early, and higher LSR can indicate hesitation rather than useful scanning.
Human demonstrations have lower ECR ($0.60$) because they terminate upon finding the target, whereas failed model episodes may run until the step budget is exhausted.
The opening $12$-step panoramic sweep is excluded from LSR.

\section{Model Architecture Details}
\label{app:architecture}

\ourmethod is built on InternVL3-2B (InternViT $300\,$M + Qwen2 $1.5\,$B, connected by an MLP bridge with GELU).
The $448{\times}448$ BEV input yields $256$ tokens after pixel shuffle (downsample ratio $0.5$).
We extend the base model with dual visual routing, spatial embedding modules, and reserved candidate-ID tokens.

\paragraph{Dual-ViT routing.}
The BEV map and ego RGB memories have markedly different visual statistics, so we process them with two separate ViT branches, both initialized from pretrained InternViT and adapted with independent LoRA modules.
The BEV crop receives $256$ tokens; each ego image receives $32$ tokens (pooled from $256$ via group-of-$8$ averaging), keeping the per-candidate token cost compact.

\paragraph{Spatial injection.}
For BEV visual tokens, a \texttt{PositionEmbedding2D} produces a $1536$-d embedding ($768$-d sinusoidal position + $768$-d learned heading projection) added at the agent cell and rendered candidate pixels.
For text tokens, the \texttt{<s>} token receives an absolute pose embedding via \texttt{text\_pos\_mlp}. Each candidate's relative-geometry feature $\mathbf{f}_i$ (Eq.~\ref{eq:rel-feat}) is passed through a \texttt{PairwiseSpatialEncoder} ($3{\to}384{\to}1536{\to}1536$, GELU) and a dedicated \texttt{cand\_pos\_mlp} ($1536{\to}1536{\to}1536$), then added to \texttt{<cand>} and, when dual text-position injection is enabled, also to \texttt{<e\_cand>}. Separate adapters for agent-state and candidate-geometry embeddings avoid sharing projections across different input distributions.

\paragraph{Candidate-ID special tokens.}
Thirty-two reserved tokens \texttt{<id\_0>}--\texttt{<id\_31>} are added to the vocabulary. Their embeddings are initialized from the pretrained embedding mean and updated via a gradient hook while the rest of the LLM embedding table stays frozen. The angular soft loss is computed over the active ID token set at the answer position.

\paragraph{Module-level training policy.}
Table~\ref{tab:trainable_modules} details the training policy. The pretrained BEV ViT, ego ViT, and Qwen2 LLM are frozen and adapted with LoRA (rank $64$ for BEV ViT and LLM, rank $16$ for ego ViT). The MLP bridge and spatial modules are fully trainable. Candidate-ID embeddings are updated via gradient hook. The model is fine-tuned for $3$ epochs on $4$ A100 GPUs.

\begin{table}[h]
\centering
\small
\caption{\textbf{Training policy per module.} LoRA is applied to the frozen pretrained visual and language backbones; newly introduced projection and spatial modules are fully trainable.}
\label{tab:trainable_modules}
\begin{tabular}{lll}
\toprule
Module & Base weights & Adaptation \\
\midrule
BEV ViT & Frozen & LoRA rank $64$ \\
Ego ViT & Frozen & LoRA rank $16$ \\
MLP bridge (\texttt{mlp1}) & Fully trainable & --- \\
LLM & Frozen & LoRA rank $64$ \\
\texttt{PositionEmbedding2D} & Fully trainable & --- \\
\texttt{text\_pos\_mlp} & Fully trainable & --- \\
\texttt{PairwiseSpatialEncoder} & Fully trainable & --- \\
\texttt{cand\_pos\_mlp} & Fully trainable & --- \\
\texttt{<id\_k>} token embeddings & Trainable via gradient hook & --- \\
\bottomrule
\end{tabular}
\end{table}

\section{Training Corpus Statistics}
\label{app:training-corpus}

\ourmethod is trained on the Habitat-Web ObjectNav demonstrations~\citep{ramrakhya2022habitat}, collected via AMT in MP3D~\citep{chang2017matterport3d} scans.
Crowd workers navigated in first-person RGB view without depth or GPS+Compass, producing demonstrations averaging ${\sim}243$ steps with ${\sim}88.9\%$ SR and ${\sim}39.9\%$ SPL.
The full dataset contains ${\sim}70$k demonstrations across $56$ scenes ($\sim$19.5M steps), exhibiting substantially more exploration than oracle shortest-path trajectories.

\paragraph{Scene and episode selection.}
We use $28$ MP3D training scenes overlapping with the standard ObjectNav split, retaining $21$ goal categories after excluding fine-grained/ambiguous ones (e.g., \texttt{foodstuff}, \texttt{fruit}, \texttt{plaything}). $23{,}767$ episodes are loaded for replay.

\paragraph{Replay filtering.}
We remove invalid/overly long replays, discard \texttt{look\_up}/\texttt{look\_down} actions (fixed horizontal camera), and remove idle frames. Episodes are replayed with $640{\times}480$ RGB-D, $79^\circ$ FoV, $0.88\,\mathrm{m}$ camera height.

\paragraph{Panoramic initialization.}
A $12$-step in-place panoramic sweep ($30^\circ$ per \texttt{turn\_right}) is prepended to initialize the BEV map and candidate visual memories.

\paragraph{Episode-level quality filtering.}
We discard episodes where \texttt{target\_found} is not set at the final step, where the target position is missing or far from the agent, or where the final agent position is near the crop boundary.

\paragraph{Corpus summary.}
Table~\ref{tab:corpus_stats} summarizes the resulting training corpus.
After replay, filtering, candidate construction, and Frontier-based Human-Intent Labeling, the dataset contains $2{,}363{,}108$ candidate-level training samples.

\begin{table}[h]
\centering
\small
\caption{\textbf{Training corpus statistics} after replay, filtering, candidate construction, and label recovery.}
\label{tab:corpus_stats}
\begin{tabular}{lr}
\toprule
Statistic & Value \\
\midrule
Source dataset & Habitat-Web~\citep{ramrakhya2022habitat} (MP3D ObjectNav) \\
MP3D training scenes & $28$ \\
ObjectNav goal categories & $21$ \\
Loaded demonstration episodes & $23{,}767$ \\
Candidate-level training samples & $2{,}363{,}108$ \\
Average candidates per step & ${\sim}18$ \\
Maximum candidates per step & $32$ \\
\bottomrule
\end{tabular}
\end{table}

\section{Hyperparameters and Optimization}
\label{app:hyperparameters}

Table~\ref{tab:hyperparams} lists the hyperparameters used for the released \ourmethod checkpoint.
All ablation variants described in Sec.~\ref{sec:experiments} share these settings unless the specific component under study is changed.
Because we use packed training, the number of optimizer steps per epoch depends on the effective number of packed samples per device; the released checkpoint is trained for $3$ epochs on $4$ A100 GPUs.

\newpage

\begin{table}[h]
\centering
\small
\caption{\textbf{Hyperparameters for the released \ourmethod checkpoint.}}
\label{tab:hyperparams}
\begin{tabular}{ll}
\toprule
Hyperparameter & Value \\
\midrule
\multicolumn{2}{l}{\emph{Model}} \\
Base model & InternVL3-2B \\
LLM LoRA rank & $64$ \\
BEV ViT LoRA rank & $64$ \\
Ego ViT LoRA rank & $16$ \\
BEV image resolution & $448\times448$ \\
Ego image resolution & $224\times224$ \\
BEV tokens per image & $256$ \\
Ego tokens per image & $32$ (pooled from $256$) \\
Maximum candidates per step & $32$ \\
Maximum sequence length & $24{,}576$ \\
\midrule
\multicolumn{2}{l}{\emph{Optimization}} \\
Optimizer & AdamW \\
Learning rate & $1\times10^{-4}$ \\
Weight decay & $0.01$ \\
LR schedule & Cosine decay \\
Warmup ratio & $0.03$ \\
Max gradient norm & $1.0$ \\
Batch size per device & $1$ \\
Gradient accumulation steps & $1$ \\
Training epochs & $3$ \\
Precision & BF16 \\
Distributed training & DeepSpeed ZeRO Stage~2 \\
Random seed & $42$ \\
\midrule
\multicolumn{2}{l}{\emph{Intent-Aligned Objective}} \\
Angular bandwidth $\sigma$ & $25^\circ\ (0.436\,\mathrm{rad})$ \\
Loss blend $\lambda$ & $0.3$ \\
Target-safe hard CE & Enabled \\
\midrule
\multicolumn{2}{l}{\emph{Compute}} \\
GPU type & A100 (80\,GB) \\
Number of GPUs & $4$ \\
\bottomrule
\end{tabular}
\end{table}

\section{Computational Cost and Deployment Latency}
\label{app:cost}

This section reports runtime cost measured on the wheeled Mecanum robot over $20$ real-world episodes on the deployment laptop (RTX 4090 Laptop GPU, $16\,\mathrm{GB}$ VRAM). The system runs as asynchronous ROS~2 nodes: LiDAR/SLAM updates the BEV map, the camera node projects RGB images, the detector runs in parallel, and the VLM node periodically queries the policy. The quadruped and humanoid platforms have lower locomotion speed, so the overlap argument is even more favorable on those embodiments.

\paragraph{Real-robot decision pipeline.}
The VLM is triggered when the robot comes within $0.5\,\mathrm{m}$ of the current waypoint, or when a $5\,\mathrm{s}$ periodic timer expires, whichever comes first.
The detector runs asynchronously; the local planner continues executing the current waypoint during VLM inference, so decision latency overlaps with the final approach motion.

\paragraph{Per-module latency breakdown.}
Table~\ref{tab:latency} reports the per-module latency breakdown averaged over $20$ episodes.

\begin{table}[h]
\centering
\small
\caption{
\textbf{Per-module latency breakdown on the RTX 4090 Laptop GPU, averaged over $20$ real-world episodes.}
Values are mean $\pm$ std.
The VLM latency excludes the first cold-start inference; the detector and camera pipelines run asynchronously and do not block waypoint execution.
}
\label{tab:latency}
\begin{tabular}{ll}
\toprule
Module / statistic & Value \\
\midrule
LiDAR BEV map update & $53.7 \pm 10.6\,\mathrm{ms}$ \\
Frontier extraction + filtering & $68.6 \pm 20.0\,\mathrm{ms}$ \\
BEV RGB rendering & $3.0 \pm 0.7\,\mathrm{ms}$ \\
VLM inference & $1279.1 \pm 148.3\,\mathrm{ms}$ \\
\midrule
\textbf{Total decision cycle} & $1404.4 \pm 182.6\,\mathrm{ms}$ \\
\midrule
Target detector (YOLOE, parallel) & $11.5 \pm 4.6\,\mathrm{ms}$ \\
Camera decode + projection (parallel) & $2.5 \pm 1.3\,\mathrm{ms}$ \\
Camera pipeline total (parallel) & $3.7 \pm 1.8\,\mathrm{ms}$ \\
\midrule
Peak GPU memory & $5.58\,\mathrm{GB}$ \\
LiDAR scan frequency & $3.35 \pm 0.27\,\mathrm{Hz}$ \\
Robot average speed & $0.27 \pm 0.19\,\mathrm{m/s}$ \\
Waypoint reached threshold & $0.5\,\mathrm{m}$ \\
\bottomrule
\end{tabular}
\end{table}

\paragraph{Scaling with candidate count.}
VLM latency scales approximately linearly with the number of candidates (each candidate adds one ego-ViT encoding and sequence length):
\begin{equation}
\mathrm{Latency}_{\mathrm{VLM}}
\approx
59.7\,\mathrm{ms}\times |\mathcal{C}_t| + 712\,\mathrm{ms}.
\end{equation}
Below $1\,\mathrm{s}$ for $3$--$4$ candidates, ${\sim}1.2\,\mathrm{s}$ for $8$, and ${\sim}2.6\,\mathrm{s}$ for $32$.

\paragraph{Online deployment feasibility.}
The next VLM query begins when the robot is $0.5\,\mathrm{m}$ from the current waypoint. At the average speed of $0.27\,\mathrm{m/s}$, the remaining approach time is ${\sim}1.85\,\mathrm{s}$, which exceeds the decision cycle ($1404.4 \pm 182.6\,\mathrm{ms}$), so the VLM typically finishes before the robot arrives.

\paragraph{Measurement caveats.}
Absolute latency depends on power state, thermal conditions, and candidate count. Cold-start measurements (including CUDA/TensorRT warmup) are excluded from steady-state numbers.

\section{Real-World Trajectory Walkthroughs}
\label{app:realworld-traj}

\begin{figure}[t]
\centering
\includegraphics[width=0.8\textwidth]{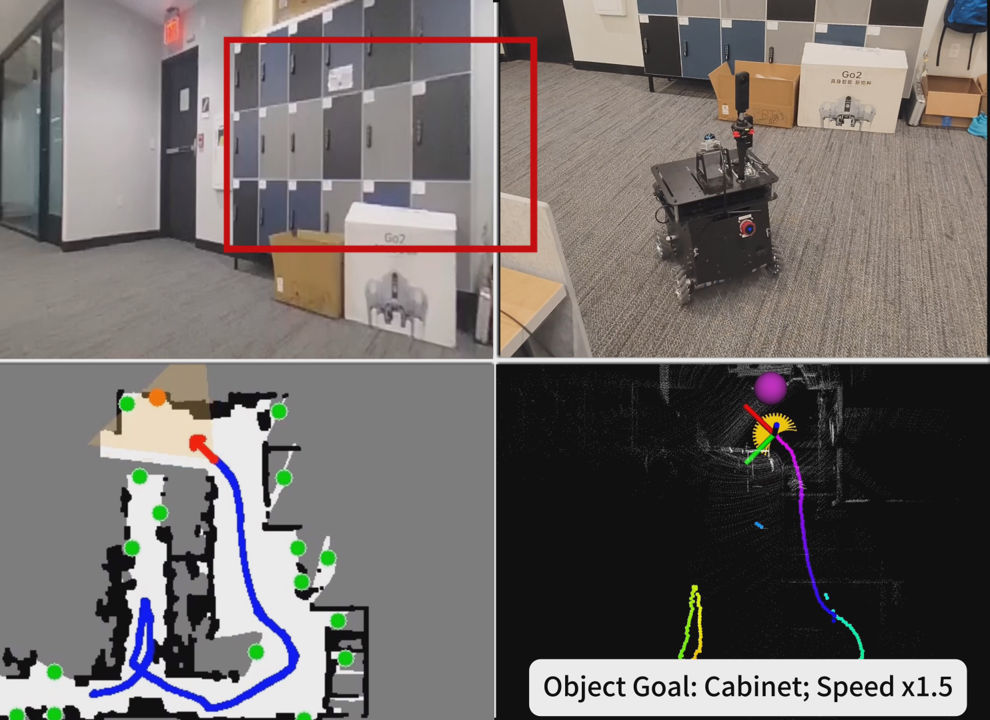}
\caption{\textbf{Wheeled robot, object goal: cabinet.} The agent navigates through a corridor lined with metal lockers. The BEV map (bottom-left) reveals that the agent initially makes a brief exploratory detour into a side passage before promptly reversing and committing to the main corridor. This early backtracking illustrates the policy's ability to quickly abandon unproductive directions and refocus exploration toward the target.}
\label{fig:real-1}
\end{figure}

\begin{figure}[h]
\centering
\includegraphics[width=0.8\textwidth]{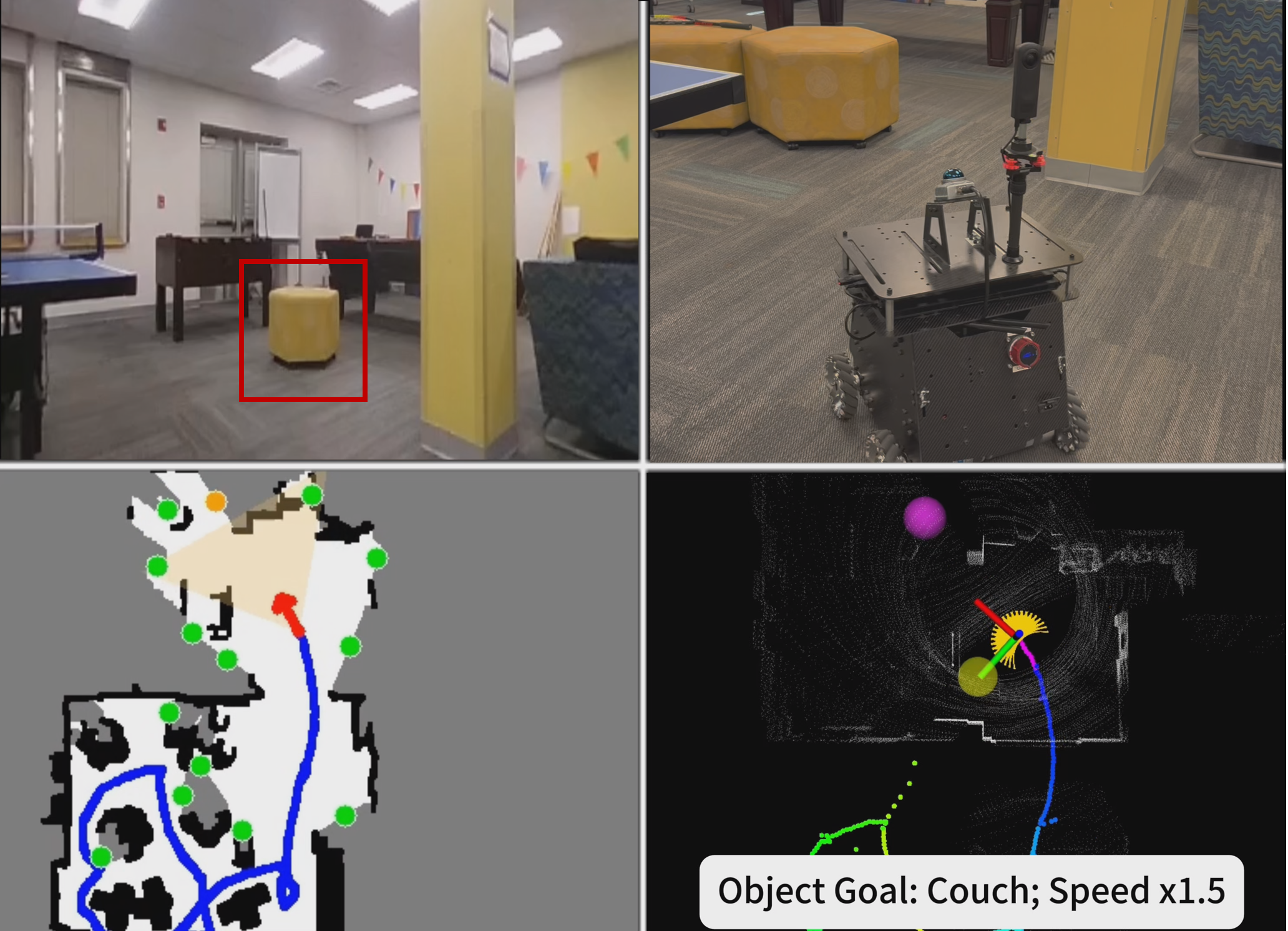}
\caption{\textbf{Wheeled robot, object goal: couch.} The agent begins in a cluttered lounge area surrounded by a ping-pong table, a foosball table, and modular seating. The BEV map (bottom-left) shows a dense initial environment with many obstacles, yet the agent efficiently sweeps through the space, exits the cluttered zone, and redirects exploration toward more promising frontiers. The policy then correctly identifies the target couch (red box, top-left) from among multiple visually similar distractors and approaches it across the open floor.}
\label{fig:real-2}
\end{figure}

\begin{figure}[t]
\centering
\includegraphics[width=0.8\textwidth]{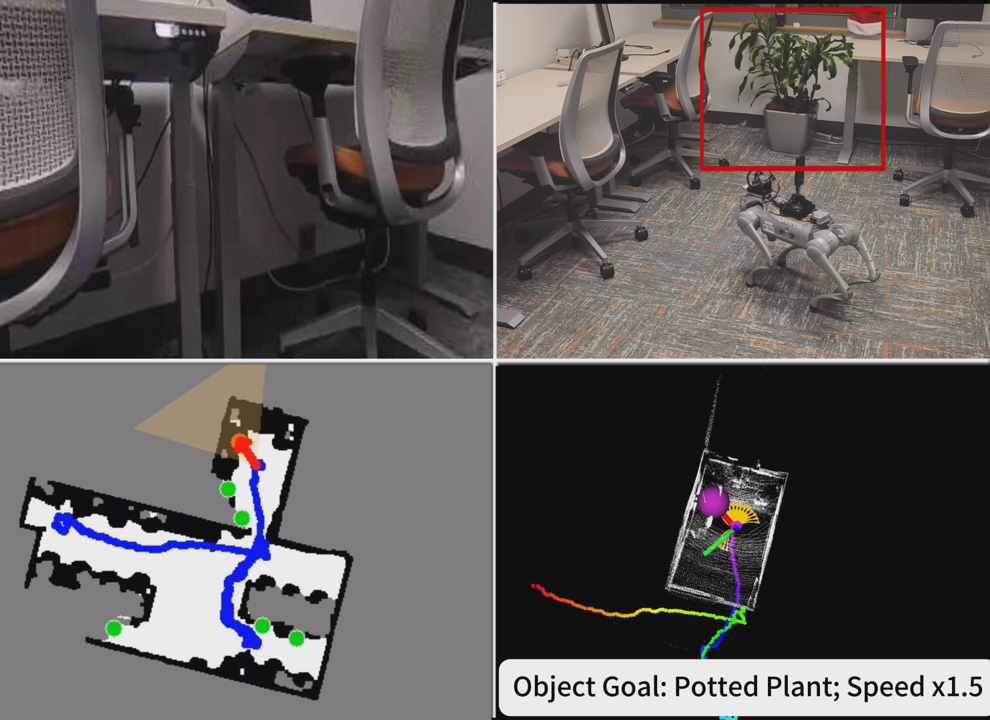}
\caption{\textbf{Unitree Go2 quadruped, object goal: potted plant.} The SLAM map (bottom-left) reveals a multi-room exploration pattern: the agent first sweeps the outer office area, and after finding no target there, proceeds into the inner room where the potted plant is located. This episode demonstrates the policy's ability to conduct structured, room-by-room search in a multi-room environment.}
\label{fig:real-5}
\end{figure}

\begin{figure}[h]
\centering
\includegraphics[width=0.8\textwidth]{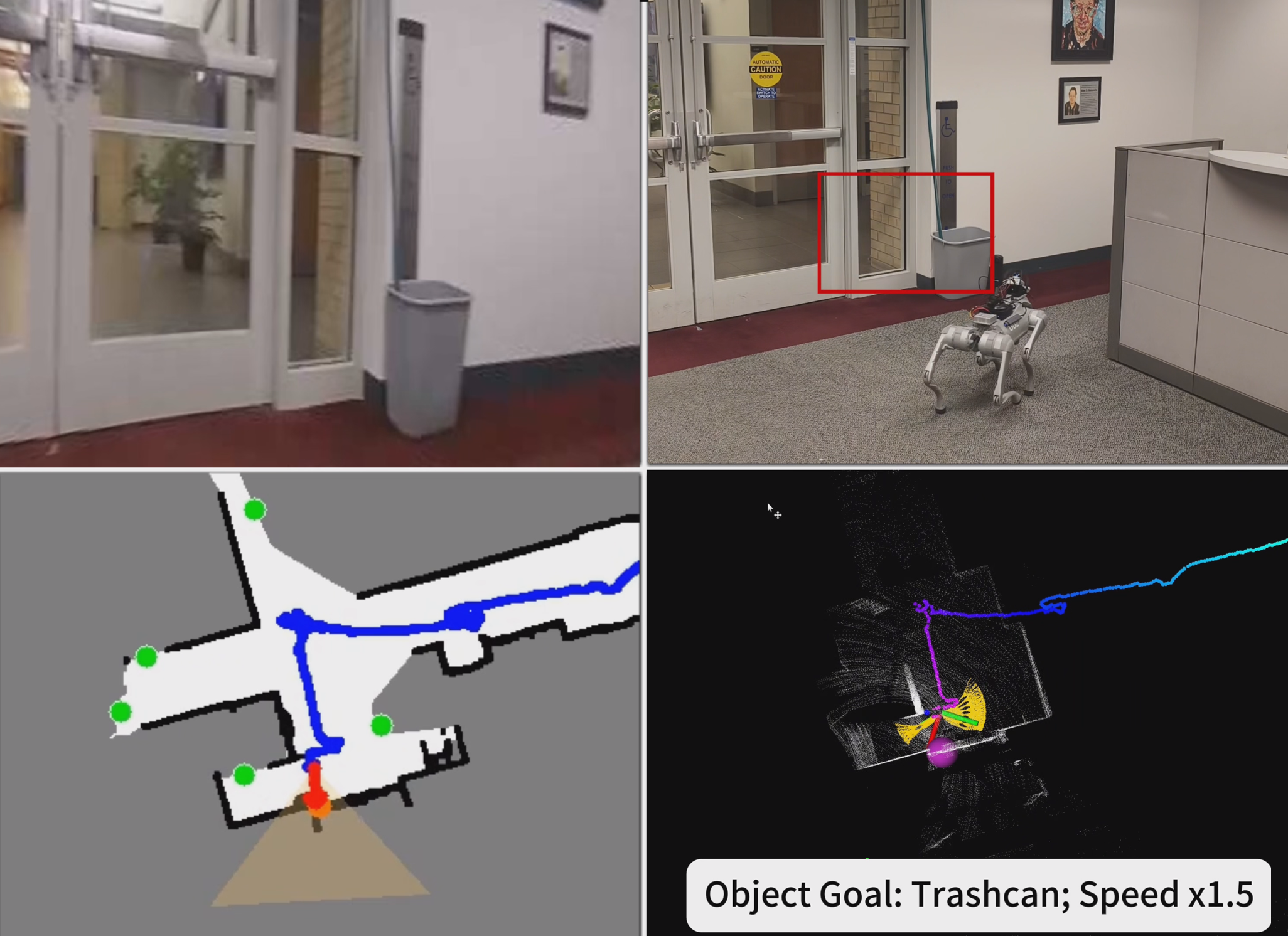}
\caption{\textbf{Unitree Go2 quadruped, object goal: trashcan.} The trajectory on the SLAM map (bottom-left) is notably long, spanning a large portion of the building across multiple corridors. Despite the extensive traversal distance, the agent's decisions are always based on the local BEV crop, which only considers nearby frontiers. This episode highlights that the local BEV abstraction scales well to large environments: the agent never needs to reason about the full global map, keeping each decision step efficient regardless of the total explored area.}
\label{fig:real-6}
\end{figure}

\begin{figure}[t]
\centering
\includegraphics[width=0.8\textwidth]{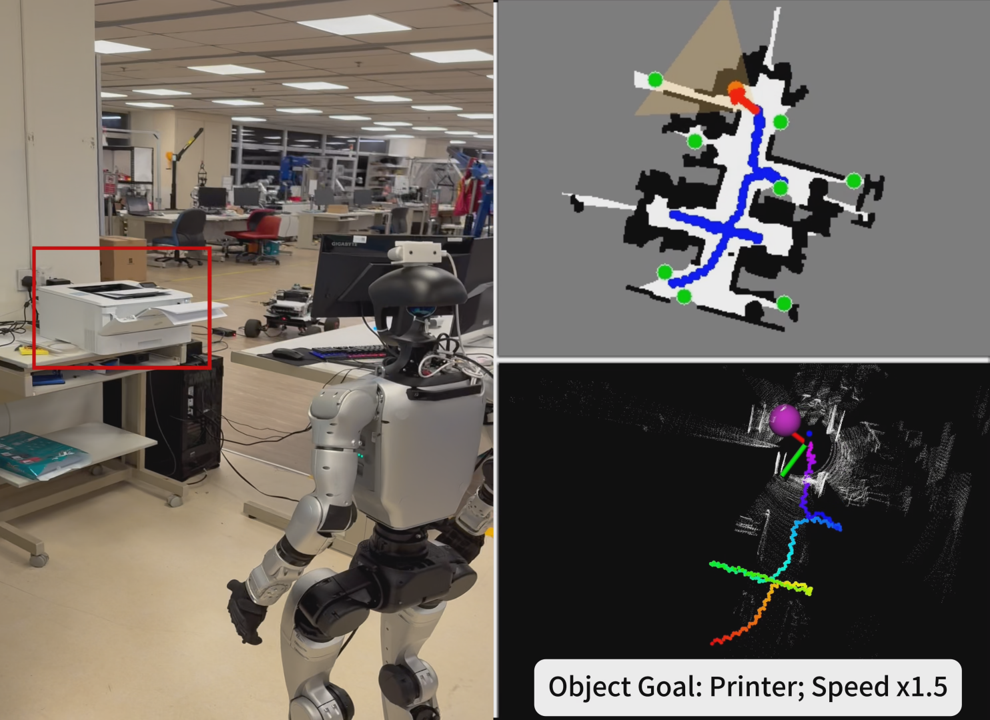}
\caption{\textbf{Unitree G1 humanoid, object goal: printer.} The humanoid navigates a densely furnished office workspace. The trajectory on the SLAM map (bottom-left) shows the agent systematically probing multiple side corridors between workstations before reversing each time, demonstrating a methodical search pattern. After exhausting several dead-end branches, the policy commits to the correct direction and locates the printer on a rolling desk. This episode illustrates that the same MP3D-trained checkpoint can be deployed through the shared waypoint interface on a bipedal platform.}
\label{fig:real-3}
\end{figure}

\begin{figure}[h]
\centering
\includegraphics[width=0.8\textwidth]{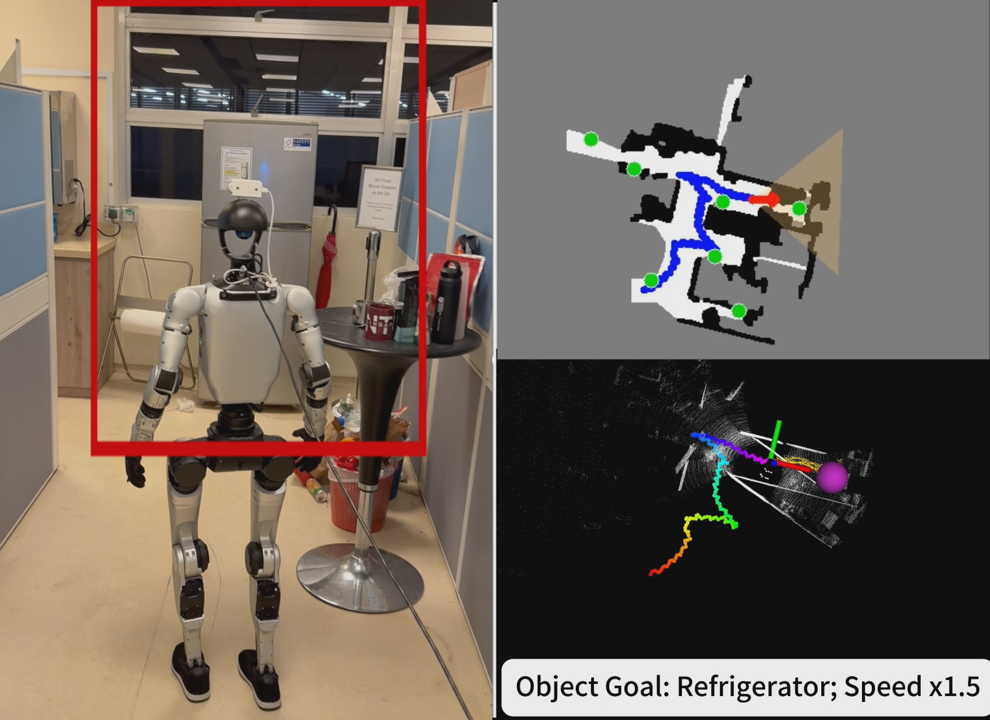}
\caption{\textbf{Unitree G1 humanoid, object goal: refrigerator.} The left panel shows the final frame with the refrigerator visible behind the robot after approach. The trajectory on the SLAM map (top-right) shows the agent first explored multiple directions in the cubicle area before committing to the target. This behavior resembles human-like search: the agent systematically surveys the environment and immediately commits to the goal once the detector confirms it.}
\label{fig:real-4}
\end{figure}

We deploy the same MP3D-trained checkpoint on three physically distinct platforms---a wheeled Mecanum robot, a Unitree Go2 quadruped, and a Unitree G1 humanoid---without any platform-specific fine-tuning. Figures~\ref{fig:real-1}--\ref{fig:real-4} present representative trajectories. Despite differences in locomotion, sensing, and embodiment morphology, the policy produces consistent exploration and target-approach behavior across all three platforms.

\newpage

\newpage


\newpage


\end{document}